\pdfoutput=1
\documentclass{article}

\PassOptionsToPackage{numbers, compress}{natbib}

\usepackage[eandd, final]{neurips_2026}

\usepackage[utf8]{inputenc} %
\usepackage[T1]{fontenc}    %
\usepackage{hyperref}       %
\usepackage{url}            %
\usepackage{booktabs}       %
\usepackage{amsfonts}       %
\usepackage{nicefrac}       %
\usepackage{microtype}      %
\usepackage{xcolor}         %
\usepackage{graphicx}
\usepackage{enumitem}
\usepackage{listings}       %
\usepackage{fancyvrb}       %
\usepackage{tikz}           %
\usetikzlibrary{positioning, arrows.meta, calc, fit, shapes.geometric}
\usepackage{subcaption}     %
\usepackage{amsmath}
\usepackage{microtype}
\usepackage{wrapfig}
\usepackage{needspace}

\definecolor{shopcodebg}{HTML}{F5F5F7}
\definecolor{shopcoderule}{HTML}{C8C8CE}
\lstdefinestyle{shopguru-base}{%
    basicstyle=\scriptsize\ttfamily,
    breaklines=true,
    breakatwhitespace=true,
    backgroundcolor=\color{shopcodebg},
    frame=single,
    framerule=0.4pt,
    rulecolor=\color{shopcoderule},
    framesep=5pt,
    xleftmargin=7pt,
    xrightmargin=7pt,
    aboveskip=6pt,
    belowskip=6pt,
    showstringspaces=false,
    columns=fullflexible,
    frame=lines,
    framesep=3pt,
    aboveskip=4pt,
    belowskip=4pt,
    captionpos=b,
    upquote=true,
}
\lstdefinelanguage{shopjson}{%
    morestring=[b]",
    keywords={true,false,null},
    keywordstyle=\color{purple!70!black}\bfseries,
    stringstyle=\color{teal!70!black},
    sensitive=true,
}
\lstdefinestyle{shopjson}{%
    style=shopguru-base,
    language=shopjson,
}
\lstdefinestyle{shopprompt}{%
    style=shopguru-base,
    literate={→}{{->}}2 {€}{{EUR}}3
             {š}{{s}}1 {Š}{{S}}1
             {ė}{{e}}1 {ą}{{a}}1
             {į}{{i}}1 {ų}{{u}}1
}

\title{ShopGym: An Integrated Framework for Realistic Simulation and Scalable Benchmarking of E-Commerce Web Agents}

\author{%
  Chinmay Savadikar$^{1,2,\dagger,}$\thanks{This work was done as part of internship at Shopify}\ , 
  Mingyu Zhao$^{2,}$\thanks{Equal contribution, order decided randomly}\ ,
  Yuanzheng Zhu$^{2,\dagger}$,
  Han Li$^{2}$,
  Shuang Xie$^{2}$,\\
  \textbf{Alberto Castelo}$^{2}$,
  \textbf{Tianfu Wu}$^{1}$,
  \textbf{Lingyun Wang}$^{2,}$\thanks{Corresponding author: \texttt{lingyun.wang@shopify.com}}\\[0.5em]
  $^{1}$North Carolina State University \qquad
  $^{2}$Shopify \\
  {\tt Website}: \url{https://shopgym-research.github.io/}
}

\begin{document}

\maketitle

\begin{abstract}
Developing and evaluating e-commerce web agents requires environments that preserve meaningful task structure while enabling controllable, reproducible, and scalable scientific comparison. Existing methodologies force a tradeoff: live storefronts provide realism but are non-stationary, difficult to inspect, and irreproducible, while hand-built sandbox benchmarks provide control but cover only a narrow range of layouts, catalogs, policies, and interaction patterns. We argue that the core bottleneck is methodological: the field lacks a scalable way to construct evaluation settings that are simultaneously realistic, diverse, controllable, inspectable, and reproducible. We introduce ShopGym, an integrated framework for realistic simulation and scalable benchmarking of e-commerce web agents. ShopGym is a framework for constructing e-commerce simulation environments and grounded benchmark tasks. Its simulation layer, ShopArena, converts live seed storefronts into self-contained sandbox shops through anonymized shop specifications and a staged, validated generation process. On top of these simulated storefronts, ShopGuru synthesizes benchmark tasks across seven skill categories, grounding each task in the shop's catalog, navigation structure, policies, and interaction affordances. Together, ShopArena and ShopGuru produce self-contained, resettable, inspectable, and stable evaluation artifacts that preserve structural properties and agent-evaluation signals relevant to shopping tasks. We validate the framework through graph-based structural analysis and agent-based behavioral evaluation with 224 generated tasks across six sandbox shops: three constructed with synthetic data and three with real data. Our results show that the synthetic shops preserve key structural properties of live storefronts, and that agent success rates on synthetic twins closely track those on live storefronts, with matching per-skill profiles and failure modes. Generating ten shops shows that the pipeline scales at about \$60 per shop with declining human effort.

\end{abstract}

\section{Introduction}

As web agents become more capable, the field increasingly needs stable environments that support long-horizon interaction in realistic, user-facing settings. E-commerce is a particularly demanding domain for this purpose: shopping tasks require agents to navigate multi-page interfaces, interpret product information, satisfy user constraints, compare alternatives, and execute actions such as selecting variants or adding items to cart. These tasks combine perception, reasoning, and sequential decision-making in a grounded setting, making storefronts a compelling testbed for web-agent evaluation. The importance of this domain is reflected in growing interest in shopping assistants \cite{shopping-companion,lassa,product-research}, offline A/B testing \cite{simgym,agentab}, and user behavior simulation for e-commerce applications \cite{lu2025can,shopping-companion,zhang2025see,customerr1}.

Existing environments for e-commerce agents face a fundamental tradeoff between realism and experimental control. Live-websites \cite{online-mind2web,webvoyager,deepshop} offer strong realism: agents are run and evaluated against realistic layouts, and naturally occurring variation and interaction patterns. But live storefronts are inherently non-stationary. Catalogs, layouts, and webpage design on live storefronts can change, collections can be reorganized, and storefronts participating in live A/B testing can serve separate variants for different requests. This non-stationary behavior makes reproducible training and evaluation of web agents challenging. The evaluation performance can reflect not only agent capability but also incidental variation from website drift, location-specific behavior, and operational noise, making results  difficult to reproduce and agents difficult to compare fairly \cite{online-mind2web}, and training runs difficult to reproduce.

Sandbox environments \cite{webarena,visualwebarena,webshop,shoppingbench,shopsimulator,webmall}, in contrast, are controllable and repeatable by design, but at the cost of realism and diversity of live websites. Manually constructing realistic storefront environments is expensive and covers only a narrow range of layouts, catalog structures, and interaction patterns. 

Existing work proposing environments for e-commerce agents has largely explored the two methodologies, with works like DeepShop \cite{deepshop} using live websites, and WebShop \cite{webshop}, ShoppingBench \cite{shoppingbench}, ShopSimulator \cite{shopsimulator}, and WebMall \cite{webmall} manually building sandbox environments. \textit{We argue that the deeper bottleneck is methodological}: the field lacks a scalable way to \emph{\textbf{construct} reliable simulated environments that are realistic, controllable, inspectable, and reproducible}.

To address this gap, we propose \textbf{ShopGym}, a framework for constructing realistic e-commerce simulation environments and grounded web-agent benchmarks. ShopGym comprises two complementary components in a synergistic workflow: (a) the simulation environment layer, \textbf{ShopArena}, transforms one or more live seed storefronts into a self-contained sandbox shop; and (b) \textbf{ShopGuru} synthesizes benchmark tasks grounded in the sandbox shops. 

\begin{figure}[t]
    \centering
    \includegraphics[width=0.98\linewidth]{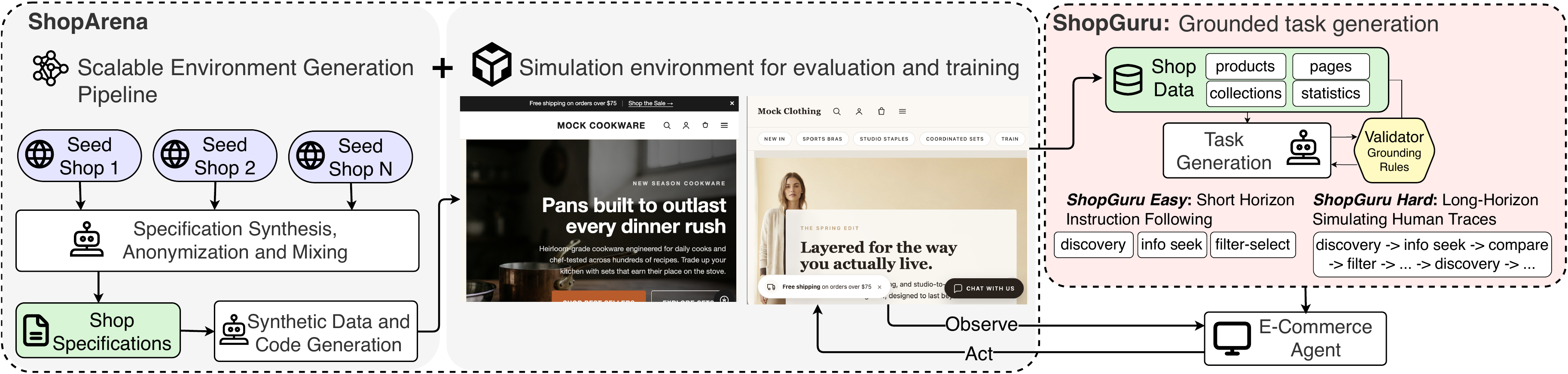}
    \caption{
    \textbf{ShopGym} comprises two components. \textbf{ShopArena} provides a simulation environment populated with synthetic sandbox shops, along with a scalable pipeline that generates new sandbox shops from one or more live seed storefronts through specification synthesis followed by data and code generation. \textbf{ShopGuru} then consumes the resulting catalog, collections, pages, and shop statistics to generate both short-horizon tasks covering primitive skills and long-horizon shopping journeys that combine these skills.
    }
    \vspace{-5mm}
    \label{fig:full-pipeline}
\end{figure}

\textbf{ShopArena: validated simulation environment.} To generate the simulation environment grounded in real storefronts, ShopArena separates storefront understanding (exploration) from sandbox synthesis (generation) through an intermediate specification document that serves as the sole interface between exploration and generation. Given one or more seed storefronts, ShopArena autonomously explores the sites and writes an anonymized specification capturing their navigation structure, user-facing affordances, high-level page organization, and catalog statistics. When multiple seeds are provided, ShopArena composes their structural and behavioral signals into a single specification, so one sandbox shop can span diversity that no individual source storefront would cover. A staged generation pipeline then synthesizes a runnable sandbox shop from the specification, iteratively refining the implementation through build checks, type checking, and multimodal visual validation. ShopArena does not aim to clone a particular storefront; it produces \textit{behaviorally aligned approximations} -- synthetic shops grounded in real storefront structure and interaction patterns, but with synthetic brand identity, products, descriptions, and imagery. The goal is not visual or brand-level fidelity but structural, behavioral, and statistical alignment sufficient for controlled evaluation. \emph{Because the specification is human-readable, benchmark creators can also edit generated environments in controlled ways without re-exploring the source storefronts}.

\textbf{ShopGuru: grounded task generation.} ShopGuru consumes the sandbox shop's catalog, navigation graph, and store policies and synthesizes tasks that are valid against the environment they will be executed in. This design separates live-web exploration from benchmark execution. The generated shops are stable and resettable, while the intermediate specification provides an inspectable control surface that benchmark creators can edit without re-exploring the source storefronts. Generating many shops does not reintroduce live-web non-stationarity: ShopGym adds \emph{controlled} variation \emph{across} frozen, resettable, versioned environments, rather than uncontrolled drift within one.

We evaluate ShopGym by generating synthetic sandbox shops and anonymized twin shops, then validating them using structural metrics and agent-based behavioral comparisons. Across paired tasks defined over source and generated shops, agent success rates on sandbox environments closely track success rates on live storefronts with matching per-skill profiles and failure modes, indicating that generated storefronts retain meaningful evaluative signal from live-web settings while substantially improving control and reproducibility. We further measure the cost, human effort, and diversity of generating ten shops. While it does not completely replace  live-web evaluation, ShopGym provides a stronger methodology for standardized benchmarking and analysis. To summarize, our contributions are:

\begin{itemize}[leftmargin=*]
    \item We identify the realism-control tradeoff in e-commerce environments and agent evaluation as a methodological bottleneck, and argue  that the missing capability is scalable construction of realistic, controllable, and reproducible environments and evaluation benchmarks.
    \item We bridge this gap with \textbf{ShopGym}, a framework for constructing realistic, controlable, and reproducible e-commerce agent evaluation environments.
    \item We present \textbf{ShopArena}, a pipeline that converts live seed storefronts into anonymized, self-contained sandbox shops through specification synthesis and staged code generation.
    \item We present \textbf{ShopGuru}, a grounded task-generation pipeline that created short-horizon and long-horizon shopping tasks tied to each generated shop's catalog, navigation structure, filters, and policies.
    \item We validate that the simulated environments preserve structural properties and meaningful signal from live storefront evaluation, supporting their use as controlled, behaviorally aligned benchmarking environments, and we measure the cost, stability, and diversity of generating shops at scale.
\end{itemize}

\section{ShopGym}
ShopGym combines two complementary frameworks: \textbf{ShopArena}, a simulation environment composed of interactable sandbox shops, and \textbf{ShopGuru}, an evaluation framework with a synthetic task generation pipeline grounded in the sandbox shops. ShopArena consists of a simulation environment containing multiple sandbox shops. It also provides a scalable pipeline to extend the environment by generating new synthetic sandbox shops: given one or more seed storefronts $\{S\}_{N}$ (real e-commerce websites), the ShopArena generation pipeline can generate a stable, self-contained, and anonymized sandbox shop $E$ using a multi-agent shop exploration and code synthesis pipeline. The ShopGuru evaluation framework generates a set of instruction following tasks $T$ defined over the sandbox shop $E$. Following standard practice~\cite{webarena, gym-anything}, we define a task $T = (E, {s_0}, p, V)$ as a sandbox shop $E$ with initial state $s_0$, a natural-language intent $p$, and a verification function $V$ that maps the agent's trajectory to a binary success score. Figure \ref{fig:full-pipeline} illustrates the full pipeline.

\subsection{ShopArena}
\label{sec:shop-arena}
To generate each sandbox shop in the ShopArena environment, we propose a scalable pipeline that converts one or more live seed storefronts $\{S\}_N$ into a realistic, self-contained, inspectable and reproducible sandbox shop. \emph{Importantly, the pipeline can be extended to any live storefront, making ShopArena an evolving environment that can be updated with new sandbox stores}.

\textbf{The generation pipeline}. Doing this end-to-end in one pass is an extremely long-horizon task: it combines storefront exploration and understanding, synthetic catalog construction, and source-code generation in a single trajectory that is hard to control and debug, and prone to incomplete outputs as agentic LLMs accumulate context~\cite{lost-in-the-middle, illusion-of-diminishing-returns}. We instead split synthesis into two phases connected by a single intermediate artifact: an anonymized design specification $M$. The \emph{exploration} phase (\S\ref{sec:shop-explore}) browses $\{S\}_N$ and writes $M$; the \emph{generation} phase (\S\ref{sec:shop-gen}) reads $M$ and writes the source code for a runnable sandbox shop. This split confines all live-web dependence to exploration, lets either phase be re-run independently, and turns the human-readable specification into a control surface for editing the generated environment without re-exploring the source.

\textbf{Multi-agent framework.} Both phases are organized as small sets of agents communicating through the file system. Each agent is an instance of a coding agent differentiated by (a)~the tools it is given access to and (b)~the objective described by its system prompt. Agents share a workspace directory but do not share a conversational context. The harness and model are replaceable: the shops in \S\ref{sec:sandbox-shop-validation} use Claude Code with Claude Opus 4.6, and we also run the pipeline with the open-source Pi harness driving GPT-5.5 or GLM-5.2 (\S\ref{sec:scalability}). Appendix~\ref{app:pipeline-invocations} classifies every pipeline stage as deterministic, a direct LLM call, or a coding-agent session.

\subsubsection{Exploration: From Live Storefront to Specifications}
\label{sec:shop-explore}

\begin{figure}[t]
    \centering
    \includegraphics[width=0.8\linewidth]{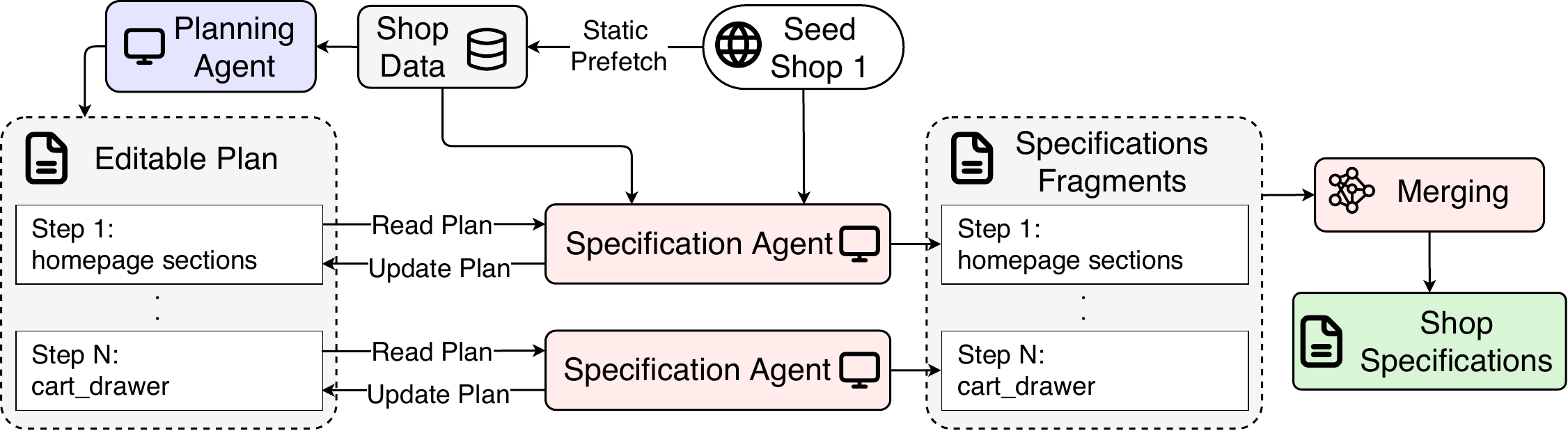}
    \caption{The exploration phase fetches key pages, and the public product catalogue. A planner agent then writes a plan using a list of predefined features to implement. Finally, Specification agents write anonymized specifications for each sub-task, which are merged to form the shop specifications.}
    \label{fig:shop-explore}
\end{figure}

The exploration phase browses one or more seed storefronts $\{S\}_N$, extracts the properties of a live storefront that matter for downstream environment generation, and generates a specification document $M$. The specification consists of three facets: (i) a design manual, which is a natural language description of the store and its visual style; (ii) a structured list of its attributes that are typically present on storefronts such as filter attributes, navigation menu design, pagination information, etc. (iii) statistics such as number of products, price range and distribution, etc. Figure \ref{fig:shop-explore} shows the full exploration pipeline.

\textbf{Static prefetch}. The phase begins with a deterministic prefetch of canonical surfaces (homepage HTML, sitemap, search and cart endpoints, public catalog JSON, policy pages).

\textbf{Planner agent ($\textsc{Agent}_\text{plan}$).} A single planner consumes the prefetched files and, with a small live-browsing budget, writes a plan that decomposes exploration of $S_i$ into focused subtasks. The planner is shown an example plan but is required to adapt it to $S_i$, so coverage is grounded in what the storefront actually exposes rather than in a fixed template. An example plan is shown in Fig \ref{fig:shoparena-plan-md}.

\textbf{Specification agent ($\textsc{Agent}_\text{spec}$).} For each planned subtask, the harness spawns a fresh specification agent equipped with browser-automation tools (Playwright). The agent performs targeted browsing, captures evidence (screenshots, snapshots, optional XHR traces), and writes an anonymized fragment of $M$ for its subtask. $\textsc{Agent}_\text{spec}$ is instructed to ignore brand names, product names, named individuals, and absolute URLs at write time, so everything downstream of exploration is anonymous by construction --- no separate redaction pass is required. Since generation sees only $M$, any residual source-specific information must appear in the human-readable $M$, an explicit checkpoint for human review.

\textbf{Consolidation and multi-seed composition.} A non-agentic LLM merges the per-subtask fragments into one specification per feature, yielding $M$. For a sandbox shop calibrated on multiple seeds, exploration runs independently on each storefront, producing one $M(S_i)$ per seed; a second LLM merge composes them into a single composite specification. Multi-seed composition is how ShopArena scales \emph{within} a shop: a single sandbox can span structural and behavioral diversity that no individual source storefront would cover.

At the end of the exploration phase, the pipeline produces the anonymized shop specifications $M$ for the sandbox shop, with the design document divided into multiple predefined \emph{features} (website skeleton, homepage, collections, etc.) A full list of features and an example of the full specifications is provided in Appendix \ref{sec:specifications-example}

\subsubsection{Generation: From Specifications to Sandbox Shop}
\label{sec:shop-gen}

The generation phase synthesizes a runnable storefront from $M$. Because it sees only $M$ and never the prefetched evidence, generation is anonymous by construction. The generation phase has two parts: a catalog synthesis pipeline and a source-code synthesis loop.

\textbf{Synthetic catalog generation.} The generation step first synthesizes the collection details (description, associated products), product names, product descriptions, and product images using the design manual and statistics from $M$. The data synthesis is structured as simple LLM calls that generate the data for one collection at a time, followed by schema checks. Store and vendor names are drawn from a configurable allowlist screened for collisions with existing brands (Appendix~\ref{app:brand-allowlist}). The details of the models used for the data generation are provided in Appendix \ref{sec:model-details}.

\textbf{Stepwise source-code synthesis.} ShopArena generates the storefront source code in a fixed sequence of \emph{steps}, each scoped to one feature: site shell and homepage, collection pages, product detail page, cart, search and filtering, policy pages, and a final integration pass. Each step is given only the slice of $M$ relevant to its feature, the catalog, and the current state of the codebase, and is executed by a dedicated execution--verification loop. Steps run in a fixed order over a shared workspace, so later steps build on a stable surface produced by earlier ones, and a failure in one step can be repaired without regenerating the rest of the shop.

\begin{figure}[t]
    \centering
    \includegraphics[width=0.8\linewidth]{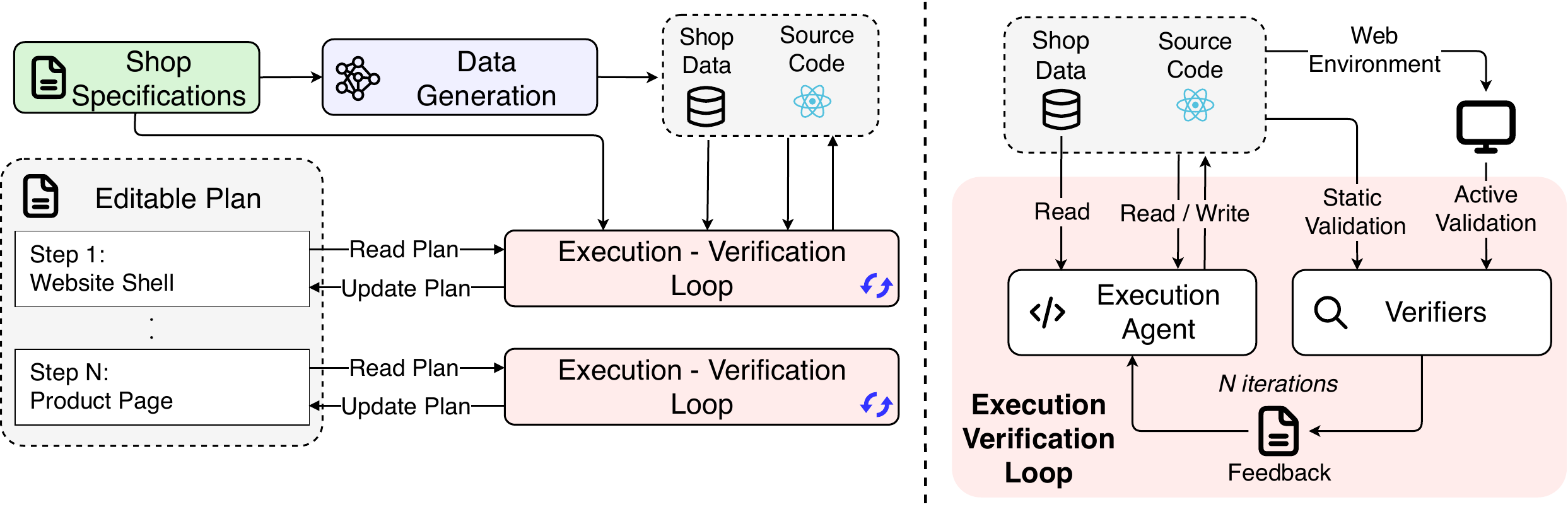}
    \caption{The generation phase. A staged sequence of feature-scoped steps is each driven by an execution--verification loop in which a fresh Execution Agent edits the codebase and a separate Verification Agent writes natural-language feedback.}
    \label{fig:generation-pipeline}
\end{figure}

\textbf{Execution--verification loop.} Within each step, code generation is structured in the spirit of the Ralph technique\footnote{\url{https://ghuntley.com/ralph/}}: a fresh agent process is spawned at each iteration, reads the current state of the workspace from disk, edits the codebase, and exits; the harness then runs verification, writes feedback to disk, and spawns the next iteration. Two specialised agents alternate within the loop:

\begin{itemize}[leftmargin=1.2em,itemsep=2pt,topsep=3pt]
\item \textbf{Execution Agent ($\textsc{Agent}_\text{exec}$).} At each iteration, a coding agent with the codebase and terminal access receives the feature to implement, the slice of $M$, the catalog, and (if present) the verifier's most recent feedback. It edits the source code and runs lightweight checks (type checking, build).
\item \textbf{Verifiers.} The code generated by $\textsc{Agent}_\text{exec}$ is checked by both rule-based verifiers and a multimodal verification agent. The rule-based verifiers ensure the site builds successfully, required pages return valid HTTP success codes, and outputs are properly structured. The verification agent then hosts the shop locally, uses browser automation to inspect the rendered site for visual and functional issues, and produces natural-language feedback for the next $\textsc{Agent}_\text{exec}$ iteration.
\end{itemize}

\textbf{Long-context management.} Each step could in principle be driven by one long-running agent that holds the codebase, the verifier's reports, and the slice of $M$ in a single context window. We avoid this because long-horizon code generation degrades as context grows~\cite{lost-in-the-middle, illusion-of-diminishing-returns}, and we observe the same failure mode in our own development: partially broken outputs the agent itself reports as complete. Spawning a fresh process at each iteration externalises three concerns that would otherwise pile up inside one prompt: the \emph{state context} (the current source code, read directly from disk), the \emph{implementation context} (the agent's prior tool-use traces, discarded), and the \emph{evaluation context} (the verifier's most recent feedback, at a known path). Each fresh $\textsc{Agent}_\text{exec}$ iteration sees only what the workspace and the latest verifier note convey; the prompt does not grow unboundedly, and the agent re-establishes context from the artifacts that matter. The loop runs for a fixed iteration budget (30 by default) or terminates early when the agent reports no further action is required. This leaves headroom: shops in \S\ref{sec:scalability} need $17$--$26$ iterations.

\textbf{Final consolidation.} After all feature-specific steps complete, ShopArena runs a final consolidation pass: a global agent navigates a tour of canonical routes, and fixes the remaining defects raised by the verifiers in prior steps.

\subsection{ShopGuru: Grounded Task Generation}
\label{sec:shopguru}

\begin{figure}[t]
    \centering
    \includegraphics[width=0.8\linewidth]{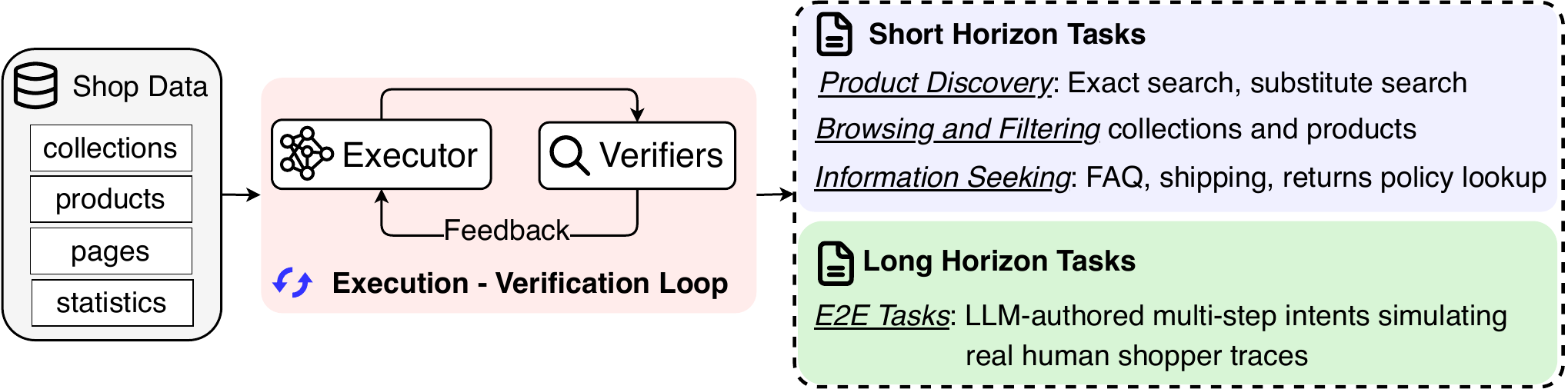}
    \caption{The ShopGuru task generation pipeline. Deterministic generators emit short-horizon tasks (\texttt{search-exact}, \texttt{search-substitute}, \texttt{browse}, \texttt{filter}, \texttt{shipping}, \texttt{returns}); an LLM-authored generator emits long-horizon shopping journeys reconciled against the shop data via a validator-driven polish loop. All emitted tasks are checked by rule-based verifiers (Table~\ref{tab:shopguru-rules}).}
    \label{fig:shopguru-dataflow}
\end{figure}

ShopGuru uses the collections, product details, and statistics from ShopArena's generation phase (\S\ref{sec:shop-gen}) and generates synthetic tasks grounded in the sandbox shop. The pipeline uses a small set of \emph{deterministic generators}, which construct short horizon tasks meant for evaluating primitive skills, and a single \emph{LLM-authored} generator for long-horizon shopping journeys. The pipeline is structured as an execution-verification loop: the generated tasks (using rule-based or LLM-based generators) are verified using multiple rule-based verifiers. The verifiers validate that the collections, products, filter types and options, and pages referenced by the task are present in the shop's data. A detailed description of the verifiers can be found in Table \ref{tab:shopguru-rules}. Figure~\ref{fig:shopguru-dataflow}
illustrates the full pipeline.

\subsubsection{Skill Catalog}
\label{sec:shopguru-skills}

We organize ShopGuru's seven skill categories into three short-horizon groups and one long-horizon group, summarized in Table~\ref{tab:shopguru-skills} in Appendix~\ref{app:shopguru-skills}. The three short-horizon groups are themselves the underlying primitives that the long-horizon journeys re-combine.

\subsubsection{Short-Horizon Task Generation}
\label{sec:shopguru-deterministic}

Short-Horizon tasks are meant to evaluate the basic instruction following capabilities of agents. The short-horizon tasks are generated using rule-based generators to evaluate three primitive skills:

\textbf{Product Discovery (\texttt{search-exact}, \texttt{search-substitute}).}
The discovery generators sample products from the catalog and emit two
task variants per product: one asking the agent to find the product by
its exact title, and one asking for a semantically similar
alternative. To keep the dataset diverse, the sampler accepts at most
one product per \texttt{product\_type}, and excludes inactive products,
gift cards, and products with no available variant (so an agent is
never asked to add an out-of-stock item).

\textbf{Filter-Selection (\texttt{browse}, \texttt{filter}).}
Browse tasks pick a populated, non-generic collection (at least three
products; generic catchall handles like \texttt{all}, \texttt{sale},
\texttt{featured}, or \texttt{best-sellers} excluded) and ask the
agent to navigate to the collection page and add any product to cart.
The filter generator additionally pairs the collection with a
realistic facet $(\textit{dim}, \textit{value})$. The non-trivial
design decision here is feasibility: shop-wide option statistics are
a poor source for per-collection filter facets because a dimension
can be common in the catalog overall yet absent from a specific
collection. We therefore build a \emph{per-collection} option index
from the collection's own products and only emit a task when at least
one product in the collection actually has the chosen
$(\textit{dim}, \textit{value})$ pair. As a fallback, when no priority
variant-option dimension is feasible in a collection, we use the
universal product fields \texttt{Vendor} (storefront-rendered as
``Brand'') and \texttt{ProductType} (rendered as ``Type''), which
every storefront filter UI exposes regardless of variant-option
coverage.

\textbf{Information-Seeking (\texttt{shipping}, \texttt{returns}).}
The policy generator searches the shop's extracted \texttt{pages.json}
for pages whose title or handle matches a small keyword list
(\textit{shipping}/\textit{delivery} for shipping;
\textit{return}/\textit{refund}/\textit{exchange} for returns). Both
the \texttt{/pages/$\langle$handle$\rangle$} route and the platform-default
\texttt{/policies/$\langle$name$\rangle$-policy} route are emitted as
soft URL hints (\texttt{url\_contains} plus a
\texttt{url\_contains\_alt} list), since some storefronts surface the
policy under a custom page handle and others under the platform
default.

\subsubsection{LLM-Authored Long-Horizon Journeys}
\label{sec:shopguru-e2e}

Long-horizon journeys are the workhorse of the benchmark's
\texttt{hard\_long\_horizon} bundle. They mimic the kind of multi-step
behavior we observe in real human shopper traces --- filter
$\rightarrow$ sort $\rightarrow$ inspect $\rightarrow$ detour to a
policy page $\rightarrow$ add to cart $\rightarrow$ edit
quantity --- which the primitive-task generators by themselves do not
cover. Authoring such journeys by hand is prohibitively expensive at
 scale, so ShopGuru delegates the long-horizon
authoring to an LLM (e.g.\ GPT-5 or Gemini 3 Pro) with two structural
guards: (i) a few-shot prompt grounded in the shop's actual catalog,
navigation, and policy pages, and (ii) a polish loop that uses the
same validator from \S\ref{sec:shopguru-validate} as a feedback
signal.

Listing~\ref{lst:e2e-task} shows a representative end-to-end (E2E) task that chains
collection navigation, a Color filter, a price sort option, and a
final cart action. The few-shot examples in the user prompt template
(eight in total; the full prompt and three representative examples are
in Appendix~\ref{app:e2e-prompts}) are constructed to reflect
commonly observed click-stream patterns in human shopper sessions on production
storefronts, drawing on prior characterizations of online shopping
behavior \cite{opera, lu2025can}: multi-pet household shopping,
returns-cautious browse-without-purchase, free-shipping-threshold
calculation, cross-brand comparison, filter-recover, first-time-visitor
info-page detour, homepage-callout brand drilldown, and so on.

The polish loop (Appendix~\ref{app:shopguru-polish-loop}) follows from a known
weakness of single-pass LLM authoring: the model occasionally
hallucinates a product into the wrong collection, asks for a variant
option that a product does not expose (e.g.\ ``select a Color'' on a
gift card), or names a filter dimension that has no realizable values
in the target collection. We close this gap by reusing the validator
as a critic. After each generation pass, the validator returns a list
of per-task issues; ShopGuru filters to the actionable subset (every
error-severity finding plus the hallucination-class warnings) and
constructs a polish prompt that lists \emph{only the flagged tasks}
alongside their specific issues and the shop context, and asks the
LLM to regenerate \emph{only those tasks} while preserving their IDs.
The polish loop runs for at most two targeted repair rounds after the initial generation. In our experiments, this
is sufficient to drive the residual error rate to zero on most shops
without human review. When tasks still fail validation after two
rounds, the build halts with a non-zero exit code rather than
silently shipping the flawed tasks; a human curator then corrects the
flagged tasks, either by re-running generation with a different seed or hand-authoring replacements
under \texttt{data\_sources/}, which override the auto-generated
outputs in the next build.

\subsubsection{Post-Generation Validation}
\label{sec:shopguru-validate}

A small dependency-free validator runs over every emitted benchmark
file as the final gate. Its seven core rules
(Table~\ref{tab:shopguru-rules})---four error-severity and three
warning-severity---target the failure modes we observe during
multi-shop benchmark regeneration; collectively they catch every
class of grounding mismatch we encounter. An actionable subset
of the validator's findings (every error plus the two
hallucination-class warnings) is also reused as the polish signal
for the LLM-authored journey generator (\S\ref{sec:shopguru-e2e}),
so the same audit logic both gates the output and steers the LLM
toward feasible tasks; the remaining \texttt{unknown-page} warning
is advisory only.

\section{Evaluating Sandbox Shop Quality}
\label{sec:sandbox-shop-validation}
For the synthetic shops in the ShopArena environment to be a reasonable substitute for live storefronts, we want the sandbox shops to be \textit{structurally and behaviorally} similar to real websites. Structural validation aims to evaluate the static structure of the synthetic shop against the real shops, whereas behavioral validation evaluates shopping workflows.

To demonstrate and validate the ShopArena as a structurally and behaviorally aligned environment, we generate three synthetic sandbox shops spanning clothing, cookware and electronics hardware domains. We refer to these shops as {\tt sandbox shops}. The screenshots of the sandbox shops can be found in Section \ref{sec:shoparena-generation-example}, along with the links to the live hosted sandbox shops. We further generate three synthetic shops using real product data instead of synthetic data, which we refer to as {\tt twin shops}. When generating the twin shops, we use an additional visual verification agent that matches the visual theme and structure of the real storefronts, and shop-specific information such as names, locations, and contact details is replaced with mock information.

\subsection{Structural Validity}
\label{sec:structural-validity}

To validate the structure of the {\tt sandbox shops,} we compare them with real shops along two axes: \textit{Observation and Interaction Complexity}, and \textit{State Transition Complexity}. We use 7 real shops and 3 synthetic shops for the comparisons. We note that this is an unpaired validation.

\begin{figure}[t]
    \centering
    \begin{subfigure}[t]{0.42\textwidth}
      \centering
      \includegraphics[width=\linewidth]{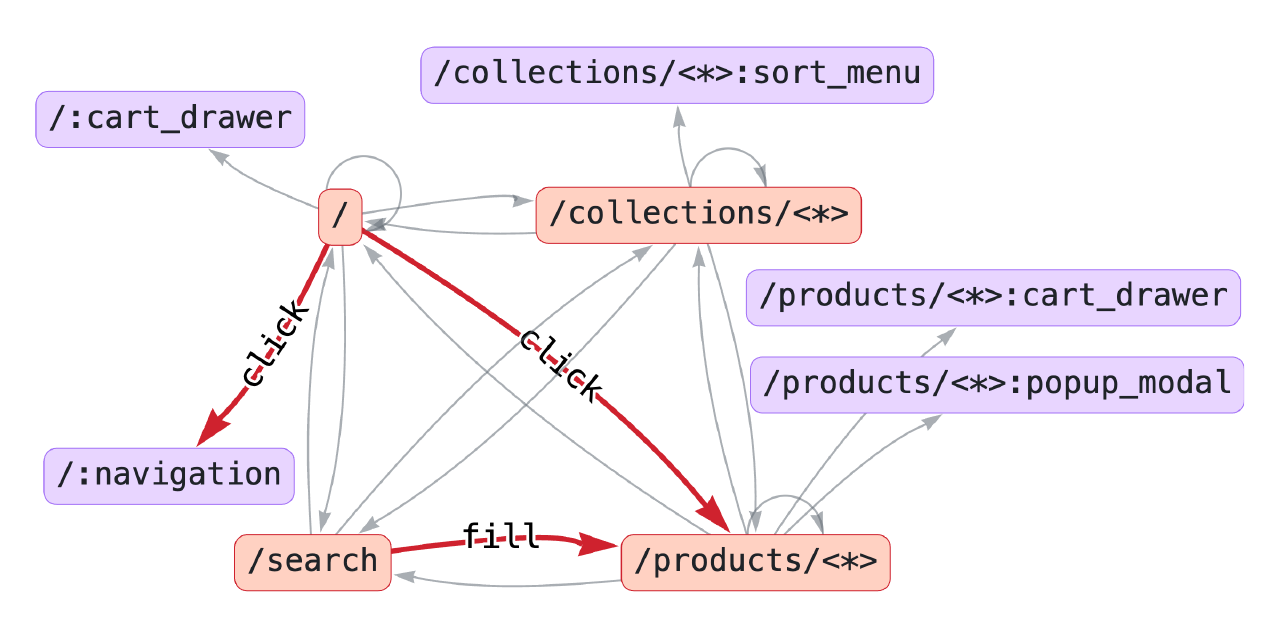}
      \caption{}
      \label{fig:graph-transition}
    \end{subfigure}
    ~
    \begin{subfigure}[t]{0.23\textwidth}
      \centering
      \includegraphics[width=\linewidth]{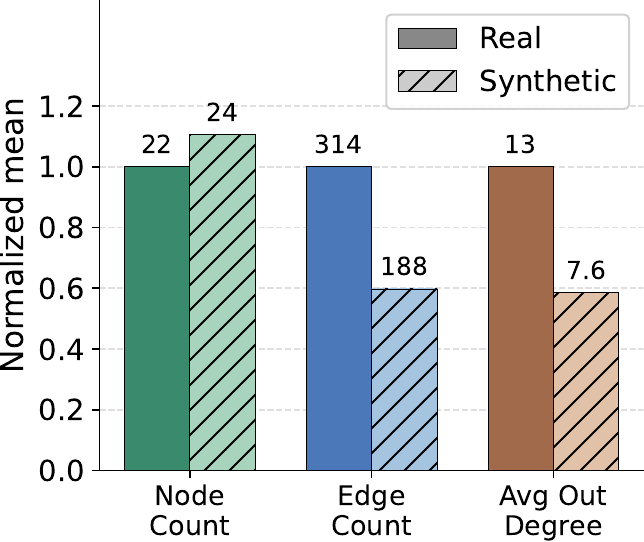}
      \caption{}
      \label{fig:result-transition-statistics}
    \end{subfigure}
    ~
    \begin{subfigure}[t]{0.28\textwidth}
      \centering
      \includegraphics[width=\linewidth]{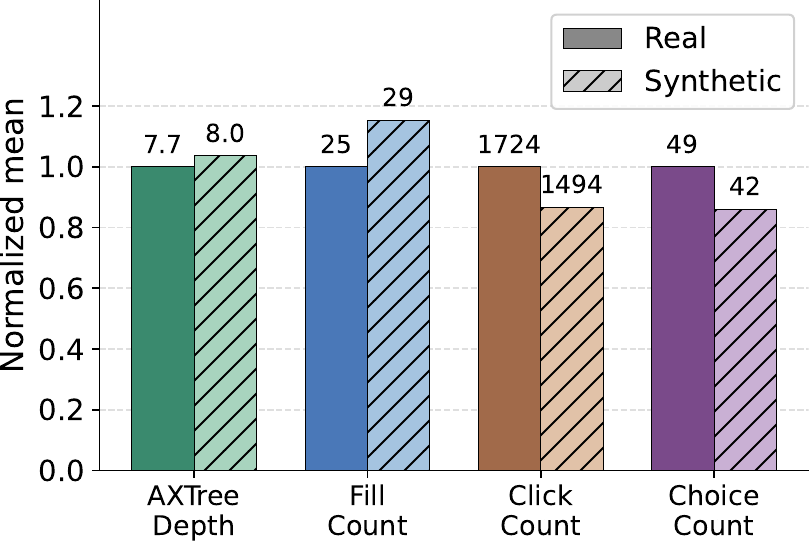}
      \caption{}
      \label{fig:obs-statistics}
    \end{subfigure}
    \caption{(a) An example directed graph visualization of the website; (b) the State Transition statistics; (c) the Observation/Interaction statistics}
    \label{fig:transition-statistics}
\end{figure}

\textbf{Observation/Interaction Complexity}. To compare the observation space, we use the depth of the simplified accessibility tree as a proxy for evaluating the nesting complexity of the websites\footnote{As defined by BrowserGym}. The nesting complexity evaluates how much structured content the agents must parse. We compare the interaction complexity by counting the number of unique elements on the website that support the {\tt fill} action, {\tt click} action, or elements that support select style controls ({\tt choice}). Figure \ref{fig:obs-statistics} shows that the synthetic sandbox shops are comparable in terms of observation and interaction complexity. In absolute terms, the real and synthetic means are $7.7$ vs.\ $8.0$ for semantic AXTree depth, $25$ vs.\ $29$ for {\tt fill} targets, $1724$ vs.\ $1494$ for {\tt click} targets, and $49$ vs.\ $42$ for {\tt choice} controls.

\textbf{State Transition Complexity}. To compare the state transition complexity, we treat the website as a directed state-transition graph, where each node corresponds to a distinct UI state, as illustrated in Figure~\ref{fig:graph-transition}. Each pink node (for example {\tt /}) denotes individual pages, and violet nodes ({\tt /:navigation}) denote alternate UI configurations of those pages when interactive elements are opened. Edges denote the actions that trigger transitions between the states, for instance, the red arrow shows clicking the navigation bar on {\tt /} transitions the state to {\tt /:navigation}. 

The aggregated statistics in Figure~\ref{fig:result-transition-statistics} show that the synthetic sandbox shops have a comparable number of nodes to the real shops ($22$ vs.\ $24$ on average), indicating similar coverage of distinct UI states. However, they exhibit fewer edges ($314$ vs.\ $188$) and a lower average out-degree ($13$ vs.\ $7.6$), meaning that each state tends to have fewer outgoing transitions. This reduction is largely because real storefronts contain additional navigation paths through external links, marketing pages, and auxiliary site components that are intentionally excluded from the sandbox environments. Overall, the synthetic shops closely match real storefronts in observation complexity, action affordances, and transition-graph scale, supporting their use as realistic sandbox environments.

As this comparison is unpaired and small, it characterizes central tendency rather than spread; per-shop ranges over ten shops are in \S\ref{sec:scalability}. Generation is also stable: across three reruns of one shop from the same specification, every structural metric varies by under $8\%$ (coefficient of variation), and LLM judges rate reruns as visually similar ($0.83$ on a $0$--$1$ scale; Appendix~\ref{app:generation-stability}).

\subsection{Behavioral Validation}
\label{sec:results-codegen}

\begin{figure}[t]
    \centering
    \begin{subfigure}[t]{0.48\textwidth}
      \centering
      \includegraphics[width=\linewidth]{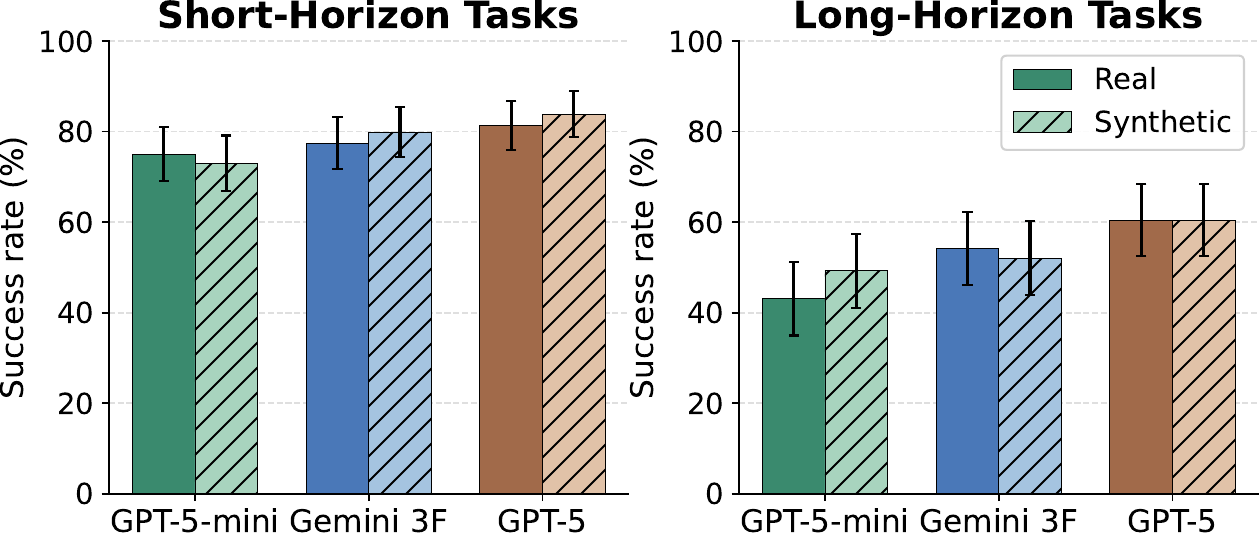}
      \caption{BrowserGym Harness.}
      \label{fig:twin-shop-results-browsergym}
    \end{subfigure}
    ~
    \begin{subfigure}[t]{0.48\textwidth}
      \centering
      \includegraphics[width=\linewidth]{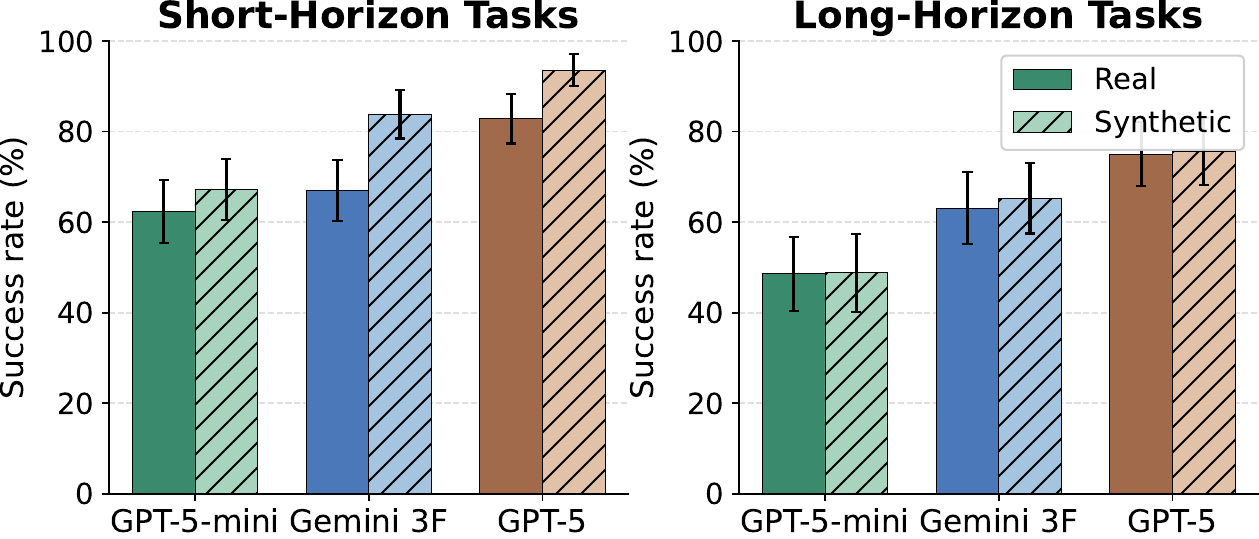}
      \caption{Internal Harness.}
      \label{fig:twin-shop-results-custom}
    \end{subfigure}
    \caption{Success rates on real storefronts vs. ShopArena twins.}
    \label{fig:twin-shop-results}
\end{figure}

We use the {\tt twin shops} to perform behavioral validation of the ShopArena pipeline. The {\tt twin shops} are visually and structurally similar to real shops, with the shop-specific information replaced with mock information. We then generate the ShopGuru evaluation suite for the real and twin shops, and evaluate multiple frontier models on both variants. We evaluate GPT-5-mini, GPT5 \cite{gpt5}, and Gemini 3 Flash.

We evaluate the models using two harnesses: a BrowserGym \cite{browsergym} based implementation that uses accessibility trees (AXTrees), and a proprietary implementation that uses screenshots along with AXTrees. We use GPT-5 as a judge in both implementations, which analyzes the agent trajectories and provides a binary success rate. The implementation details are provided in Appendix \ref{sec:web-agent-implementations}.

Figure \ref{fig:twin-shop-results} shows that all models achieve similar performance on the real storefront and its twin variant. Furthermore, the performance on the complex
tasks improves with stronger models. This shows that from the
perspective of agent behavior, the sandbox shops are behaviorally
consistent with real shops.

\textbf{Real vs.\ twin trajectories.} Similar aggregate success does not imply identical behavior. Running GPT-5 on matched real and twin variants of every task (three shops, three seeds), trajectory length ($13.1$ vs.\ $13.3$ steps), click share ($74.8\%$ vs.\ $74.2\%$), and success ($72.7\%$ vs.\ $72.4\%$) are nearly identical. Per-family success (Table~\ref{tab:real-twin-family}) is highly correlated (Pearson $r = 0.954$, Spearman $\rho = 0.964$), though filtering and policy lookup are easier on the twin, consistent with its cleaner DOM and shorter policy pages. Failure patterns also agree ($r = 0.960$): most failures are specification-compliance errors, where the agent reaches the right collection, product, or cart but misses an exact route, filter, variant, or final-response requirement. Yet $26.8\%$ of matched runs differ in outcome: twins exercise the same skills at comparable difficulty, but are not replicas.

\begin{table}[t]
\centering
\caption{Real--twin agreement for GPT-5 on matched task variants across the
three twin shops. \textbf{Left:} success rate (\%) by ShopGuru task family.
\textbf{Right:} distribution of failure patterns (\%) over failed trajectories.}
\label{tab:real-twin-family}
\label{tab:real-twin-failures}
\footnotesize
\setlength{\tabcolsep}{3pt}
\begin{minipage}[t]{0.47\linewidth}
\centering
\begin{tabular}{lccc}
\toprule
\textbf{Task family} & \textbf{Real} & \textbf{Twin} & \textbf{$\Delta$} \\
\midrule
Exact product discovery & 100.0 & 100.0 & $\phantom{+}0.0$ \\
Substitute discovery    & 88.9  & 93.3  & $+4.4$ \\
Collection browsing     & 80.0  & 77.8  & $-2.2$ \\
Filtering               & 66.7  & 75.8  & $+9.1$ \\
Multi-step (long-horizon) & 60.4 & 55.6 & $-4.8$ \\
Returns-policy lookup   & 44.4  & 66.7  & $+22.3$ \\
Shipping-policy lookup  & 11.1  & 22.2  & $+11.1$ \\
\midrule
\textbf{All tasks}      & \textbf{72.7} & \textbf{72.4} & $-0.3$ \\
\bottomrule
\end{tabular}
\end{minipage}\hfill
\begin{minipage}[t]{0.5\linewidth}
\centering
\begin{tabular}{lcc}
\toprule
\textbf{Failure pattern} & \textbf{Real} & \textbf{Twin} \\
\midrule
Missing required URL checkpoint          & 40.0 & 41.7 \\
Final-response instruction missed        & 18.9 & 15.6 \\
Item not added on product page           & 11.1 & \phantom{0}7.3 \\
Filter incorrect or not cleared          & 10.0 & 15.6 \\
Wrong route to the right content         & \phantom{0}8.9 & \phantom{0}4.2 \\
Final cart state incorrect               & \phantom{0}7.8 & \phantom{0}7.3 \\
\bottomrule
\end{tabular}
\end{minipage}
\end{table}

\section{Validating ShopGuru task complexity}
\label{sec:results-composite}

\begin{figure}[t]
    \centering
    \begin{subfigure}[t]{0.48\textwidth}
      \centering
      \includegraphics[width=\linewidth]{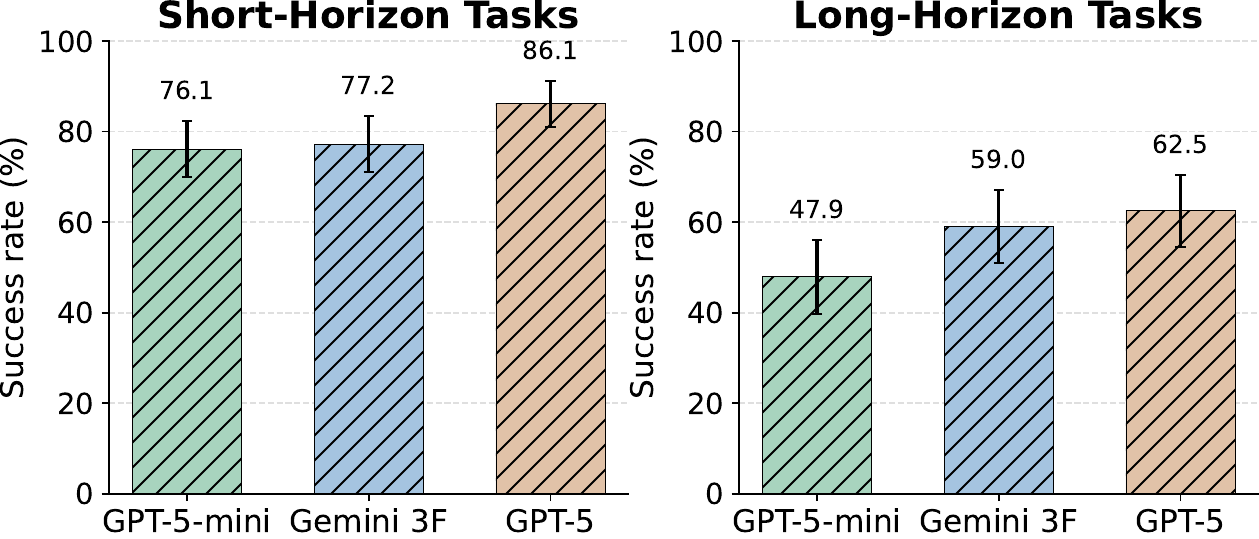}
      \caption{BrowserGym Harness.}
      \label{fig:sandbox-shop-results-browsergym}
    \end{subfigure}
    ~
    \begin{subfigure}[t]{0.48\textwidth}
      \centering
      \includegraphics[width=\linewidth]{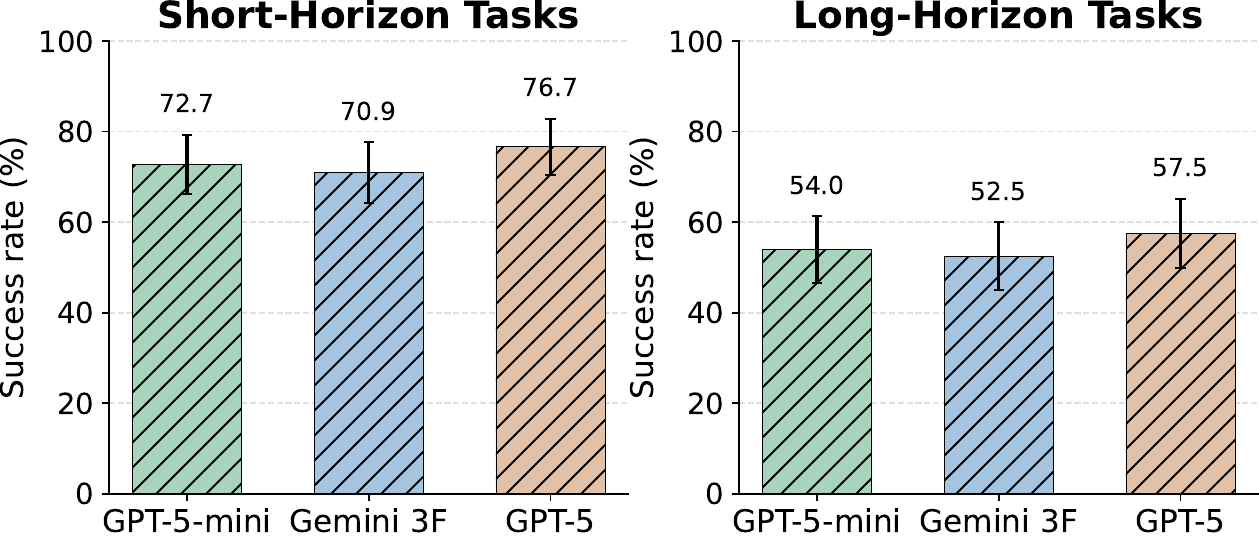}
      \caption{Internal Harness.}
      \label{fig:sandbox-shop-results-custom}
    \end{subfigure}
    \caption{Success rates on the synthetic sandbox shops.}
    \label{fig:sandbox-shop-results}
\end{figure}

To validate that the ShopGuru task generation pipeline can generate complex tasks, we evaluate frontier models with the ShopGuru evaluation suite on the {\tt sandbox shops}. We use the same web agent, harness, and judge setup as \S\ref{sec:results-codegen}.

As seen in Figure \ref{fig:sandbox-shop-results-browsergym} GPT-5-mini with the BrowserGym harness achieves a success rate of 47.9\% on the Long-Horizon complex tasks, Gemini 3 Flash achieves 59\%, and GPT-5 achieves 62.5\%. The results with the custom harness show similar performance across all models. A low performance of frontier models shows that the ShopGuru task generation pipeline can generate meaningfully hard tasks which are useful for training and evaluation of next generation e-commerce agents.

\needspace{6\baselineskip}
\section{Scalability of Environment Generation}
\label{sec:scalability}

\begin{wraptable}{r}{0.47\textwidth}
\vspace{-4mm}
\centering
\caption{Per-shop cost with Pi + GPT-5.5 (six runs over two shops).}
\label{tab:generation-cost}
\footnotesize
\setlength{\tabcolsep}{2.5pt}
\begin{tabular}{lccc}
\toprule
\textbf{Stage} & \textbf{Tokens} & \textbf{Time} & \textbf{Cost} \\
\midrule
Exploration & 8--9M   & 0.6--0.8 h & \$12--13 \\
Generation  & 17--18M & 2--3 h     & \$46--48 \\
\textbf{Total} & \textbf{25--27M} & \textbf{2.6--3.8 h} & \textbf{\$58--61} \\
\bottomrule
\end{tabular}
\vspace{-3mm}
\end{wraptable}

We generate seven additional fully synthetic shops in concurrent jobs with the open-source Pi harness and GPT-5.5. All seven produce buildable storefronts within $17$--$26$ iterations; the verifiers catch and the execution agent repairs $88$ visual/quality and $26$ functional defects, with no build failures (Appendix~\ref{app:generation-failures}). Table~\ref{tab:generation-cost} shows a per-shop cost of about \$60 and a few hours. With the open-weight GLM-5.2, shop generation costs $82\%$ less (\$8.47) at comparable structural quality, using more tokens and time (Appendix~\ref{app:glm-results}). Human inspection time falls from $60$ to $30$ to ${\sim}10$ minutes per shop across batches as recurring defects become reusable verifier checks. The shops are diverse: homepage DOM size varies sixfold and collection-page checkboxes range from $0$ to $93$, and operations such as search and filtering take different interface forms (Appendix~\ref{app:shop-diversity}). Finally, lightweight deployments (${\sim}600$\,MB vs.\ ${\sim}60$\,GB for WebArena) support parallel RL rollouts: $192$ rollouts in about $25$ minutes in a preliminary setup (Appendix~\ref{sec:infrastructure-requirements}).

\section{Conclusion}
We present ShopGym, an integrated framework for realistic simulation and scalable benchmarking of e-commerce web agents. ShopGym addresses a central methodological gap: live storefronts are realistic but non-stationary, while sandbox environments are controllable but limited in diversity. By combining ShopArena for sandbox shop generation and ShopGuru for grounded task synthesis, ShopGym produces self-contained, inspectable, resettable, and stable evaluation artifacts.

The generated shops are structurally comparable to real storefronts, and agent performance on synthetic twins closely tracks performance on live storefronts, with matching per-skill profiles and failure modes. Fully synthetic shops yield benchmarks of comparable difficulty, and generation scales at modest cost and declining human effort. ShopGym thus provides a practical middle ground between live-web realism and sandbox control, and a scalable foundation for more reproducible evaluation of e-commerce web agents.

\label{body:end}

\newpage
\bibliographystyle{plainnat}
\bibliography{main}

\begin{thebibliography}{30}
\providecommand{\natexlab}[1]{#1}
\providecommand{\url}[1]{\texttt{#1}}
\expandafter\ifx\csname urlstyle\endcsname\relax
  \providecommand{\doi}[1]{doi: #1}\else
  \providecommand{\doi}{doi: \begingroup \urlstyle{rm}\Url}\fi

\bibitem[Aggarwal et~al.(2026)Aggarwal, Neubig, and Welleck]{gym-anything}
Pranjal Aggarwal, Graham Neubig, and Sean Welleck.
\newblock Gym-anything: Turn any software into an agent environment.
\newblock \emph{arXiv preprint arXiv:2604.06126}, 2026.

\bibitem[Castelo et~al.(2026)Castelo, Foumani, Fan, Koay, Malik, Zhu, Li,
  Feghhi, Uliana, Xie, Zhang, Martins, Zhao, Pelland, Faerman, LeBlanc, Glazer,
  McNamara, Wang, and Wu]{simgym}
Alberto Castelo, Zahra~Zanjani Foumani, Ailin Fan, Keat~Yang Koay, Vibhor
  Malik, Yuanzheng Zhu, Han Li, Meysam Feghhi, Ronie Uliana, Shuang Xie, Zhaoyu
  Zhang, Angelo~Ocana Martins, Mingyu Zhao, Francis Pelland, Jonathan Faerman,
  Nikolas LeBlanc, Aaron Glazer, Andrew McNamara, Lingyun Wang, and Zhong Wu.
\newblock Simgym: Traffic-grounded browser agents for offline {A/B} testing in
  e-commerce.
\newblock \emph{CoRR}, abs/2602.01443, 2026.
\newblock \doi{10.48550/ARXIV.2602.01443}.
\newblock URL \url{https://doi.org/10.48550/arXiv.2602.01443}.

\bibitem[Chae et~al.(2026)Chae, Park, and Ritter]{verienv}
Hyungjoo Chae, Jungsoo Park, and Alan Ritter.
\newblock Safe and scalable web agent learning via recreated websites.
\newblock \emph{CoRR}, abs/2603.10505, 2026.
\newblock \doi{10.48550/ARXIV.2603.10505}.
\newblock URL \url{https://doi.org/10.48550/arXiv.2603.10505}.

\bibitem[de~Chezelles et~al.(2025)de~Chezelles, Gasse, Lacoste, Caccia, Drouin,
  Boisvert, Thakkar, Marty, Assouel, Shayegan, Jang, L{\`u}, Yoran, Kong, Xu,
  Reddy, Neubig, Cappart, Salakhutdinov, and Chapados]{browsergym}
Thibault Le~Sellier de~Chezelles, Maxime Gasse, Alexandre Lacoste, Massimo
  Caccia, Alexandre Drouin, L{\'e}o Boisvert, Megh Thakkar, Tom Marty, Rim
  Assouel, Sahar~Omidi Shayegan, Lawrence~Keunho Jang, Xing~Han L{\`u}, Ori
  Yoran, Dehan Kong, Frank~F. Xu, Siva Reddy, Graham Neubig, Quentin Cappart,
  Russ Salakhutdinov, and Nicolas Chapados.
\newblock The browsergym ecosystem for web agent research.
\newblock \emph{Transactions on Machine Learning Research}, 2025.
\newblock ISSN 2835-8856.
\newblock URL \url{https://openreview.net/forum?id=5298fKGmv3}.
\newblock Expert Certification.

\bibitem[Deng et~al.(2023)Deng, Gu, Zheng, Chen, Stevens, Wang, Sun, and
  Su]{mind2web}
Xiang Deng, Yu~Gu, Boyuan Zheng, Shijie Chen, Samual Stevens, Boshi Wang, Huan
  Sun, and Yu~Su.
\newblock Mind2web: Towards a generalist agent for the web.
\newblock In Alice Oh, Tristan Naumann, Amir Globerson, Kate Saenko, Moritz
  Hardt, and Sergey Levine, editors, \emph{Advances in Neural Information
  Processing Systems 36: Annual Conference on Neural Information Processing
  Systems 2023, NeurIPS 2023, New Orleans, LA, USA, December 10 - 16, 2023},
  2023.
\newblock URL
  \url{http://papers.nips.cc/paper\_files/paper/2023/hash/5950bf290a1570ea401bf98882128160-Abstract-Datasets\_and\_Benchmarks.html}.

\bibitem[Gandhi and Neubig(2026)]{gobrowse}
Apurva Gandhi and Graham Neubig.
\newblock Go-browse: Training web agents with structured exploration.
\newblock In \emph{The Fourteenth International Conference on Learning
  Representations}, 2026.
\newblock URL \url{https://openreview.net/forum?id=IpzRWE52yw}.

\bibitem[He et~al.(2024)He, Yao, Ma, Yu, Dai, Zhang, Lan, and Yu]{webvoyager}
Hongliang He, Wenlin Yao, Kaixin Ma, Wenhao Yu, Yong Dai, Hongming Zhang,
  Zhenzhong Lan, and Dong Yu.
\newblock Webvoyager: Building an end-to-end web agent with large multimodal
  models.
\newblock In Lun{-}Wei Ku, Andre Martins, and Vivek Srikumar, editors,
  \emph{Proceedings of the 62nd Annual Meeting of the Association for
  Computational Linguistics (Volume 1: Long Papers), {ACL} 2024, Bangkok,
  Thailand, August 11-16, 2024}, pages 6864--6890. Association for
  Computational Linguistics, 2024.
\newblock \doi{10.18653/V1/2024.ACL-LONG.371}.
\newblock URL \url{https://doi.org/10.18653/v1/2024.acl-long.371}.

\bibitem[Huang et~al.(2026)Huang, Huang, Choubey, Zhou, Chen, May, and
  Wu]{huang2026gta}
Tenghao Huang, Kung-Hsiang Huang, Prafulla~Kumar Choubey, Yilun Zhou, Muhao
  Chen, Jonathan May, and Chien-Sheng Wu.
\newblock {GTA}: Generating long-horizon tasks for web agents at scale.
\newblock In Maria Liakata, Viviane~P. Moreira, Jiajun Zhang, and David
  Jurgens, editors, \emph{Proceedings of the 64th Annual Meeting of the
  {A}ssociation for {C}omputational {L}inguistics (Volume 1: Long Papers)},
  pages 18805--18820, San Diego, California, United States, July 2026.
  Association for Computational Linguistics.
\newblock ISBN 979-8-89176-390-6.
\newblock \doi{10.18653/v1/2026.acl-long.856}.
\newblock URL \url{https://aclanthology.org/2026.acl-long.856/}.

\bibitem[Koh et~al.(2024)Koh, Lo, Jang, Duvvur, Lim, Huang, Neubig, Zhou,
  Salakhutdinov, and Fried]{visualwebarena}
Jing~Yu Koh, Robert Lo, Lawrence Jang, Vikram Duvvur, Ming~Chong Lim, Po-Yu
  Huang, Graham Neubig, Shuyan Zhou, Ruslan Salakhutdinov, and Daniel Fried.
\newblock Visualwebarena: Evaluating multimodal agents on realistic visual web
  tasks.
\newblock In \emph{Proceedings of the 62nd Annual Meeting of the Association
  for Computational Linguistics (Volume 1: Long Papers), {ACL} 2024}, 2024.
\newblock URL \url{https://arxiv.org/abs/2401.13649}.

\bibitem[Liu et~al.(2023)Liu, Lin, Hewitt, Paranjape, Bevilacqua, Petroni, and
  Liang]{lost-in-the-middle}
Nelson~F. Liu, Kevin Lin, John Hewitt, Ashwin Paranjape, Michele Bevilacqua,
  Fabio Petroni, and Percy Liang.
\newblock Lost in the middle: How language models use long contexts.
\newblock \emph{Transactions of the Association for Computational Linguistics},
  12:\penalty0 157--173, 2023.
\newblock URL \url{https://api.semanticscholar.org/CorpusID:259360665}.

\bibitem[Lu et~al.(2025)Lu, Huang, Han, Yao, Bei, Gesi, Xie, He, Wang,
  et~al.]{lu2025can}
Yuxuan Lu, Jing Huang, Yan Han, Bingsheng Yao, Sisong Bei, Jiri Gesi, Yaochen
  Xie, Qi~He, Dakuo Wang, et~al.
\newblock Can llm agents simulate multi-turn human behavior? evidence from real
  online customer behavior data.
\newblock \emph{arXiv preprint arXiv:2503.20749}, 2025.

\bibitem[Lyu et~al.(2025)Lyu, Zhang, Yan, de~Rijke, Ren, and Chen]{deepshop}
Yougang Lyu, Xiaoyu Zhang, Lingyong Yan, Maarten de~Rijke, Zhaochun Ren, and
  Xiuyi Chen.
\newblock Deepshop: A benchmark for deep research shopping agents.
\newblock \emph{ArXiv}, abs/2506.02839, 2025.
\newblock URL \url{https://api.semanticscholar.org/CorpusID:279118560}.

\bibitem[Peeters et~al.(2025)Peeters, Steiner, Schwarz, Caspary, and
  Bizer]{webmall}
Ralph Peeters, Aaron Steiner, Luca Schwarz, Julian~Yuya Caspary, and Christian
  Bizer.
\newblock Webmall--a multi-shop benchmark for evaluating web agents [technical
  report].
\newblock \emph{arXiv preprint arXiv:2508.13024}, 2025.

\bibitem[Ramrakhya et~al.(2026)Ramrakhya, Szot, Attia, Mazoure, Nguyen, Yang,
  Gan, Agrawal, and Toshev]{ramrakhya2026scaling}
Ram Ramrakhya, Andrew Szot, Omar Attia, Bogdan Mazoure, Anh Nguyen, Yuhao Yang,
  Zhe Gan, Harsh Agrawal, and Alexander Toshev.
\newblock Scaling synthetic task generation for agents via exploration.
\newblock In C.~Vondrick, B.~Hariharan, C.~Raffel, L.~Pinto, D.~Yang, and
  A.~Faust, editors, \emph{International Conference on Learning
  Representations}, volume 2026, pages 123630--123660, 2026.
\newblock URL
  \url{https://proceedings.iclr.cc/paper_files/paper/2026/file/c8a4dd7d9e13583d714ce8580da7bbc7-Paper-Conference.pdf}.

\bibitem[Singh et~al.(2025)Singh, Fry, Perelman, Tart, Ganesh, El-Kishky,
  McLaughlin, Low, Ostrow, Ananthram, et~al.]{gpt5}
Aaditya Singh, Adam Fry, Adam Perelman, Adam Tart, Adi Ganesh, Ahmed El-Kishky,
  Aidan McLaughlin, Aiden Low, AJ~Ostrow, Akhila Ananthram, et~al.
\newblock Openai gpt-5 system card.
\newblock \emph{arXiv preprint arXiv:2601.03267}, 2025.

\bibitem[Sinha et~al.(2025)Sinha, Arun, Goel, Staab, and
  Geiping]{illusion-of-diminishing-returns}
Akshit Sinha, Arvindh Arun, Shashwat Goel, Steffen Staab, and Jonas Geiping.
\newblock The illusion of diminishing returns: Measuring long horizon execution
  in llms.
\newblock \emph{ArXiv}, abs/2509.09677, 2025.
\newblock URL \url{https://api.semanticscholar.org/CorpusID:281252776}.

\bibitem[Wang et~al.(2025{\natexlab{a}})Wang, Hsu, Lu, Gu, Cui, Xie, Headean,
  Yao, Veeragouni, Liu, et~al.]{agentab}
Dakuo Wang, Ting-Yao Hsu, Yuxuan Lu, Hansu Gu, Limeng Cui, Yaochen Xie, William
  Headean, Bingsheng Yao, Akash Veeragouni, Jiapeng Liu, et~al.
\newblock Agenta/b: Automated and scalable web a/btesting with interactive llm
  agents.
\newblock \emph{arXiv preprint arXiv:2504.09723}, 2025{\natexlab{a}}.

\bibitem[Wang et~al.(2025{\natexlab{b}})Wang, Xiao, Sun, Zhao, Luo, Zhang, and
  Zeng]{shoppingbench}
Jiang Wang, Kejun Xiao, Qi~Sun, Huaipeng Zhao, Tao Luo, Jiandong Zhang, and
  Xiaoyi Zeng.
\newblock Shoppingbench: A real-world intent-grounded shopping benchmark for
  llm-based agents.
\newblock In \emph{AAAI Conference on Artificial Intelligence},
  2025{\natexlab{b}}.
\newblock URL \url{https://api.semanticscholar.org/CorpusID:280536823}.

\bibitem[Wang et~al.(2026{\natexlab{a}})Wang, Xiao, Zhao, Luo, and
  Zeng]{product-research}
Jiangyuan Wang, Kejun Xiao, Huaipeng Zhao, Tao Luo, and Xiaoyi Zeng.
\newblock Productresearch: Training e-commerce deep research agents via
  multi-agent synthetic trajectory distillation.
\newblock \emph{CoRR}, abs/2602.23716, 2026{\natexlab{a}}.
\newblock \doi{10.48550/ARXIV.2602.23716}.
\newblock URL \url{https://doi.org/10.48550/arXiv.2602.23716}.

\bibitem[Wang et~al.(2026{\natexlab{b}})Wang, Wu, Song, Wang, Chen, Li, Yan,
  Deng, Liu, Zhao, Xiong, Liu, Chen, Deng, Su, and Zheng]{shopsimulator}
Pei Wang, Yanan Wu, Xiaoshuai Song, Weixun Wang, Gengru Chen, Zhongwen Li,
  Ke~Yan, Ken Deng, Qi~Liu, Shu-Man Zhao, Shaopan Xiong, Xuepeng Liu, Xuefeng
  Chen, Wanxi Deng, Wenbo Su, and Bo~Zheng.
\newblock Shopsimulator: Evaluating and exploring rl-driven llm agent for
  shopping assistants.
\newblock \emph{ArXiv}, abs/2601.18225, 2026{\natexlab{b}}.
\newblock URL \url{https://api.semanticscholar.org/CorpusID:285050373}.

\bibitem[Wang et~al.(2025{\natexlab{c}})Wang, Lu, Li, Amini, Sun, Bart, Lyu,
  Gesi, Wang, Huang, Su, Ehsan, Alikhani, Li, Chilton, and Wang]{opera}
Ziyi Wang, Yuxuan Lu, Wenbo Li, Amir~A. Amini, Bo~Sun, Yakov Bart, Weimin Lyu,
  Jiri Gesi, Tian Wang, Jing Huang, Yu~Su, Upol Ehsan, Malihe Alikhani, Toby
  Jia-Jun Li, Lydia~B. Chilton, and Dakuo Wang.
\newblock Opera: A dataset of observation, persona, rationale, and action for
  evaluating llms on human online shopping behavior simulation.
\newblock \emph{ArXiv}, abs/2506.05606, 2025{\natexlab{c}}.
\newblock URL \url{https://api.semanticscholar.org/CorpusID:279244562}.

\bibitem[Wang et~al.(2025{\natexlab{d}})Wang, Lu, Zhang, Huang, and
  Wang]{customerr1}
Ziyi Wang, Yuxuan Lu, Yimeng Zhang, Jing Huang, and Dakuo Wang.
\newblock Customer-r1: Personalized simulation of human behaviors via rl-based
  llm agent in online shopping.
\newblock \emph{arXiv preprint arXiv:2510.07230}, 2025{\natexlab{d}}.

\bibitem[Xue et~al.(2025)Xue, Qi, Shi, Song, Gou, Song, Sun, and
  Su]{online-mind2web}
Tianci Xue, Weijian Qi, Tianneng Shi, Chan~Hee Song, Boyu Gou, Dawn Song, Huan
  Sun, and Yu~Su.
\newblock An illusion of progress? assessing the current state of web agents.
\newblock \emph{CoRR}, abs/2504.01382, 2025.
\newblock \doi{10.48550/ARXIV.2504.01382}.
\newblock URL \url{https://doi.org/10.48550/arXiv.2504.01382}.

\bibitem[Yao et~al.(2022)Yao, Chen, Yang, and Narasimhan]{webshop}
Shunyu Yao, Howard Chen, John Yang, and Karthik Narasimhan.
\newblock Webshop: Towards scalable real-world web interaction with grounded
  language agents.
\newblock In Sanmi Koyejo, S.~Mohamed, A.~Agarwal, Danielle Belgrave, K.~Cho,
  and A.~Oh, editors, \emph{Advances in Neural Information Processing Systems
  35: Annual Conference on Neural Information Processing Systems 2022, NeurIPS
  2022, New Orleans, LA, USA, November 28 - December 9, 2022}, 2022.
\newblock URL
  \url{http://papers.nips.cc/paper\_files/paper/2022/hash/82ad13ec01f9fe44c01cb91814fd7b8c-Abstract-Conference.html}.

\bibitem[Yu et~al.(2026)Yu, Xiao, Zhao, Luo, and Zeng]{shopping-companion}
Zijian Yu, Kejun Xiao, Huaipeng Zhao, Tao Luo, and Xiaoyi Zeng.
\newblock Shopping companion: {A} memory-augmented {LLM} agent for real-world
  e-commerce tasks.
\newblock \emph{CoRR}, abs/2603.14864, 2026.
\newblock \doi{10.48550/ARXIV.2603.14864}.
\newblock URL \url{https://doi.org/10.48550/arXiv.2603.14864}.

\bibitem[Yuan et~al.(2026)Yuan, Yin, Cai, and Wei]{webforge}
Peng Yuan, Yuyang Yin, Yuxuan Cai, and Zheng Wei.
\newblock Webforge: Breaking the realism-reproducibility-scalability trilemma
  in browser agent benchmark.
\newblock \emph{arXiv preprint arXiv:2604.10988}, 2026.

\bibitem[Zhang et~al.(2024)Zhang, Peng, Zhao, Hu, Zhu, Zeng, and Hu]{lassa}
Shuo Zhang, Boci Peng, Xinping Zhao, Boren Hu, Yun Zhu, Yanjia Zeng, and Xuming
  Hu.
\newblock Llasa: Large language and e-commerce shopping assistant.
\newblock \emph{CoRR}, abs/2408.02006, 2024.
\newblock \doi{10.48550/ARXIV.2408.02006}.
\newblock URL \url{https://doi.org/10.48550/arXiv.2408.02006}.

\bibitem[Zhang et~al.(2025)Zhang, Gesi, Xue, Wang, Wang, Lu, Zhan, Zeng, Cui,
  Guo, et~al.]{zhang2025see}
Yimeng Zhang, Jiri Gesi, Ran Xue, Tian Wang, Ziyi Wang, Yuxuan Lu, Sinong Zhan,
  Huimin Zeng, Qingjun Cui, Yufan Guo, et~al.
\newblock See, think, act: Online shopper behavior simulation with vlm agents.
\newblock \emph{arXiv preprint arXiv:2510.19245}, 2025.

\bibitem[Zhou(2026)]{webarena-infinity}
Shuyan Zhou.
\newblock Webarena-infinity: Generating browser environments with verifiable
  tasks at scale.
\newblock \emph{shuyanzhou.com}, March 2026.
\newblock URL \url{https://webarena.dev/webarena-infinity/}.

\bibitem[Zhou et~al.(2024)Zhou, Xu, Zhu, Zhou, Lo, Sridhar, Cheng, Ou, Bisk,
  Fried, Alon, and Neubig]{webarena}
Shuyan Zhou, Frank~F. Xu, Hao Zhu, Xuhui Zhou, Robert Lo, Abishek Sridhar,
  Xianyi Cheng, Tianyue Ou, Yonatan Bisk, Daniel Fried, Uri Alon, and Graham
  Neubig.
\newblock Webarena: {A} realistic web environment for building autonomous
  agents.
\newblock In \emph{The Twelfth International Conference on Learning
  Representations, {ICLR} 2024, Vienna, Austria, May 7-11, 2024}.
  OpenReview.net, 2024.
\newblock URL \url{https://openreview.net/forum?id=oKn9c6ytLx}.

\end{thebibliography}

\appendix

\clearpage
\section{Related Work}

\paragraph{Static web environments and benchmarks.}
Prior work on web-agent evaluation has pursued two goals: broad coverage
of real websites and reproducible executable environments. Mind2Web
\cite{mind2web} provides large-scale offline traces from 137 real
websites across 31 domains, emphasizing breadth and cross-site
generalization. OPeRA \cite{opera} similarly collects observations,
personas, rationales, and actions from real human shopping sessions on
Amazon. In contrast, WebArena \cite{webarena} provides self-hosted,
fully functional websites with executable tasks, and VisualWebArena
\cite{visualwebarena} extends this setting to visually grounded tasks.
Such environments support autonomous exploration and reproducible
evaluation, unlike purely offline datasets. Several benchmarks focus specifically on e-commerce. WebShop
\cite{webshop} offers a synthetic storefront with 1.25M products.
ShoppingBench \cite{shoppingbench} provides a large shopping sandbox
with API-based interaction, aimed primarily at intent grounding.
ShopSimulator \cite{shopsimulator} introduces a Chinese web environment
for multi-turn shopping dialog and fine-grained product
differentiation. WebMall \cite{webmall} expands to four simulated shops
with authentic product offers and comparison-shopping tasks. Although
these benchmarks enable controlled evaluation, they remain tied to a
fixed set of layouts, taxonomies, and policies, and therefore capture
only a limited slice of real-world e-commerce diversity.

\textbf{Live-website evaluation.} To address this limitation, Online-Mind2Web \cite{online-mind2web},
WebVoyager \cite{webvoyager}, and DeepShop \cite{deepshop} evaluate
agents directly on live websites, including real shopping domains.
These settings improve diversity and realism, but depend on external
sites whose layouts, inventories, flows, and access policies may change
without notice. This non-stationarity makes comparisons over time
difficult and limits reproducibility, especially for RL training.

\textbf{Automated environment and task generation.} A recent line of work uses LLMs to automate environment construction.
Gym-Anything \cite{gym-anything} treats setup as a multi-agent pipeline
that installs and configures existing software, producing CUA-World with
10K+ tasks. WebArena-Infinity \cite{webarena-infinity} generates
synthetic web apps from user manuals. WebForge \cite{webforge} uses a
Plan/Generate/Refine/Validate pipeline to create self-contained web
environments at scale, and VeriEnv \cite{verienv} constructs executable
synthetic versions of real websites with deterministic rewards exposed
through a Python SDK. These approaches scale well, but often sacrifice
fidelity: their environments are derived from manuals or ungrounded LLM
design choices and may miss the layouts, integrations, and edge cases of
production storefronts. In contrast, ShopArena grounds generation in
observed real shops: it first explores a live storefront to produce an
anonymised Shop Manual capturing taxonomy, navigation, capabilities, and
statistics, then synthesises a calibrated sandbox shop from that manual.
This preserves much more of the diversity of real storefronts while
retaining determinism and reset-ability. Go-Browse \cite{gobrowse} proposes a multi-agent pipeline for generating
tasks and trajectories by navigating websites and maintaining a graph of
visited URLs for efficient exploration.

\clearpage
\section{Implementation Details}
\label{sec:web-agent-implementations}

\subsection{Models used for agent implementations}
\label{sec:model-details}
\begin{itemize}[leftmargin=*]
    \item Planning Agent: Claude Code with Claude Opus 4.6
    \item Specification Agent: Claude Code with Claude Opus 4.6
    \item Collections generator: Claude Opus 4.6
    \item Product name and description generator: Claude Sonnet 4.6
    \item Product Image Generator: Gemini 3 Pro
    \item Execution Agent: Claude Code with Claude Opus 4.6
    \item Visual Verification Agent: Claude Code with Claude Opus 4.6, enabled with Playwright for browser automation
\end{itemize}
The shops used in \S\ref{sec:sandbox-shop-validation} are generated with the configuration above. For the additional shops in \S\ref{sec:scalability}, the coding-agent sessions are instead run with the open-source, model-agnostic Pi harness driving GPT-5.5 (or GLM-5.2, Appendix~\ref{app:glm-results}); the pipeline stages are unchanged.

\subsection{Harness A: BrowserGym Agent}
\label{app:agents-browsergym}

\textbf{Observation}. The BrowserGym harness uses the accessibility tree as the observation. We do not use screenshots for simplicity.

\textbf{Action Space}. The agent is allowed to use 20 actions:
\begin{itemize}[leftmargin=*]
        \item {\tt noop}: Do nothing, optionally waiting for the given number of milliseconds.
      \item {\tt send\_msg\_to\_user}: Send a message to the user (terminal action that ends the episode).
      \item {\tt scroll}: Scroll the page horizontally and vertically by the given pixel deltas.
      \item {\tt fill}: Type a value into an \texttt{<input>}, \texttt{<textarea>}, or \texttt{[contenteditable]} element.
      \item {\tt select\_option}: Select one or multiple options in a \texttt{<select>} element by value or label.
      \item {\tt click}: Click an element with a given mouse button and optional modifier keys.
      \item {\tt dblclick}: Double-click an element with a given mouse button and optional modifier keys.
      \item {\tt hover}: Move the mouse over an element to hover it.
      \item {\tt press}: Focus an element and press a key combination (e.g.\ \texttt{Control+Enter}, \texttt{ArrowDown}).            
      \item {\tt focus}: Move keyboard focus to the matching element.
      \item {\tt clear}: Clear the contents of an input field.
      \item {\tt drag\_and\_drop}: Drag one element and drop it onto another.
      \item {\tt upload\_file}: Click an element to open a file chooser and upload one or more files.
      \item {\tt go\_back}: Navigate back to the previous page in browser history.
      \item {\tt go\_forward}: Navigate forward to the next page in browser history.\item {\tt goto}: Navigate the current tab to the given URL.\item {\tt tab\_close}: Close the currently active tab.
      \item {\tt tab\_focus}: Activate the tab at the given index, bringing it to the front.
      \item {\tt new\_tab}: Open a new tab and make it the active one.
      \item {\tt report\_infeasible}: Declare the task impossible with a short reason (terminal action that ends the episode).
  \end{itemize}  

\textbf{Browser}. We use a headless chromium browser for all experiments with BrowserGym, with Playwright for browser interaction.

\textbf{Trajectory details}. The BrowserGym agent uses a think - action - memory loop, where only the observation from the current step is used as context. The agent is instructed to reason first using {\tt <think>...</think>} tags, output a single action in {\tt <action>...</action>} tags, and write whatever it thinks is necessary to solve the task in {\tt <memory>...</memory>} tags. The think, action and memory tags from all the steps are used as context in the future steps, along with any execution errors. The agent is allowed a maximum of 30 steps, and the episode terminates when the agent sends a message to the user using the {\tt send\_msg\_to\_user} action, reports the task as infeasible using the {\tt report\_infeasible} action, or the step threshold is reached. Episodes that reach the step budget are graded as failed.

\subsection{Harness B: Internal Harness}
\label{app:agents-internal}

The internal harness implements a single-call-per-step planner–executor
agent that runs over a textual page projection and emits one
Playwright-grounded action per step. Each task is
evaluated across several independent browser sessions to absorb
agent and storefront stochasticity.

\paragraph{Observation.}
At each step, the current page is rendered into a tree-shaped textual
projection that fuses the rendered DOM and the accessibility tree
into a single representation. Every interactive element (clickable
control, text input, combobox, menu item, etc.) is tagged with a
stable identifier of the form \texttt{[elem-N]} that the agent uses
to reference the element in its action. For all experiments in this
paper the agent additionally receives a viewport screenshot of the
current page at each step, so the observation is multimodal (page
projection plus rendered pixels).
The agent prompt also
includes the current URL and a running \emph{memory}: a per-step
textual log of prior decisions, action outcomes (success / error
strings), and the agent's own analysis. The full LLM transcript is
discarded at the end of each step --- only the memory summary, the
fresh page projection, and (optionally) the latest screenshot survive
into the next prompt --- so context length grows linearly in
\emph{step count}, not in cumulative LLM output.

\paragraph{Action space.}
Each step's LLM call is constrained by a fixed structured-output
schema with the following fields:
\begin{itemize}[leftmargin=1.4em,itemsep=2pt,topsep=3pt]
\item \texttt{analysis} --- free-form chain-of-thought over the current
state, prior memory, and the task goal.
\item \texttt{shouldEnd} --- boolean flag terminating the rollout.
\item \texttt{instruction} --- a short natural-language description
of the intended next action (e.g.\ ``click on the filter menu titled
\emph{Price}'', ``type \emph{lunch box for kids} in the header search
field'').
\item \texttt{elementId} --- the \texttt{elem-N} target of the action,
referencing one of the IDs in the current page projection.
\item \texttt{method} --- a Playwright interaction method
(\texttt{click}, \texttt{fill}, \texttt{selectOption},
\texttt{press}, \texttt{check}, etc.).
\item \texttt{arguments} --- an optional list of method arguments
(e.g.\ the value to type into a text field).
\end{itemize}
The harness translates the (\texttt{elementId}, \texttt{method},
\texttt{arguments}) tuple into a Playwright call against the live
browser session. Failures (selector misses, navigation errors,
timeouts) are surfaced back to the agent as an error string in the
next step's prompt and recorded verbatim in memory.

\paragraph{Termination and budgets.}
A rollout terminates on (i) \texttt{shouldEnd}=\texttt{true}
emitted by the agent, (ii) hitting the per-task step budget of
$40$~steps, or (iii) hitting the per-task wall-clock budget of
$850$\,s. Rollouts that exit via the step or time budget are scored
as failures regardless of the cart state at termination
(\S\ref{app:agents-judge}), since the agent itself does not declare
goal completion.

\paragraph{Browser.}
The agent runs against a remote, Playwright-compatible browser
session provisioned per task. Each session is initialised at the
task's starting URL (\texttt{*\_real} for the live source storefront
or \texttt{*\_twin} for the ShopArena sandbox), accepts a
\emph{cart-only} terminal-state convention (every shopping task in
ShopGuru ends with the agent ending the session once the cart
contains the requested items, never by clicking a Checkout button),
and is torn down at the end of the rollout.

\paragraph{LLM client and models.}
All LLM calls use an OpenAI-compatible chat-completion API with
structured-output (JSON) decoding. We evaluate three models spanning
compact and flagship capability tiers: GPT-5~mini, Gemini~3~Flash,
and GPT-5. Where the underlying model exposes it, reasoning effort is
set to \texttt{medium}. Each LLM call is retried up to three times
on transient API errors before the rollout is marked failed.

\paragraph{Repeats and aggregation.}
We run three independent rollouts per (task, model, environment) cell.
Per-task verdicts are averaged within each cell, and the error bars
in Figs.~\ref{fig:twin-shop-results}b and
\ref{fig:sandbox-shop-results}b span the per-cell standard error of
the mean across the three repeats. Per-bundle pass rates aggregate
across all tasks in the bundle (\texttt{easy\_short\_horizon} or
\texttt{hard\_long\_horizon}) and all repeats.

\subsection{LLM-as-Judge}
\label{app:agents-judge}

Both harnesses score every rollout with the same LLM-as-judge model but slightly different protocols. After each rollout, the judge LLM (GPT-5) receives:
\begin{itemize}[leftmargin=1.4em,itemsep=2pt,topsep=3pt]
\item the original task intent and the task's
\texttt{success\_criteria} JSON (URL hint plus any cart/response
constraints);
\item the agent's per-step memory log (textual summary of
decisions and action outcomes);
\item the agent's URL trajectory (every URL visited, in order);
\item the rollout's \texttt{terminationType}
(\texttt{goal\_reached}, \texttt{steps\_limit\_reached}, or
\texttt{time\_limit\_reached}).
\end{itemize}
The judge prompt instructs the model to weigh task satisfaction first,
followed by output quality, tool-use effectiveness, and recovery
from transient errors; it returns a structured JSON object with a
binary \texttt{success} field and a free-form \texttt{reasoning}
explanation. The verdict is then post-processed by a hard rule:
rollouts that do not terminate via the agent's own
\texttt{shouldEnd} are forced to \texttt{success}=\texttt{false}
regardless of the judge's opinion. This guards against a class of
spurious successes where the agent times out on a page whose final
state happens to look correct (e.g.\ the requested product is on the
page but is never added to cart) but never emits a deliberate
end-of-session signal.

With the BrowserGym harness, {\tt url\_contains} field is treated as a hard gate, where the trajectory is graded as failed if the expected URL is not present in the trajectory. The judge LLM receives the entire cart history, the agent's reasoning, action and memory traces, as well as the screenshots of the webpages in the trajectory and outputs a binary pass/fail decision along with it's reasoning.

\subsection{Infrastructure Requirements}
\label{sec:infrastructure-requirements}
We perform all evaluations using the official APIs from the model providers. To parallelize the evaluation suite, the agent harnesses are orchestrated on a standard cloud virtual machine.

\textbf{Parallel rollouts for training.} ShopGym also supports training through parallel rollouts. In a preliminary experiment, a rollout service manages isolated Playwright sessions against shop containers on separate VMs, allowing browser and shop hosting to scale independently. We run 24 concurrent rollouts across five VMs: one rollout-service VM (44 Intel Cascade Lake CPUs, 128\,GB RAM) and four shop-hosting VMs (24 AMD Rome CPUs and 32\,GB RAM each). Eight H200 GPUs, two for training and six for rollout generation, produce 192 rollouts in about 25 minutes. Compared with WebArena, each ShopGym deployment is about 600\,MB in our setup versus about 60\,GB, and uses lightweight resets rather than extensive database restoration, which reduces storage and reset overhead for parallel training. These results demonstrate systems feasibility, not comprehensive scaling or a completed RL evaluation.

\clearpage
\section{Limitations}
\label{sec:limitations}

ShopGym improves control, reproducibility, and scalability, but it is not a full substitute for live-web evaluation. By construction, ShopArena removes many sources of real-world non-stationarity, including inventory churn, personalization, A/B tests, third-party integrations, latency variation, and other operational noise. This is desirable for controlled benchmarking, but it also means that strong performance in ShopGym should be interpreted as evidence of capability under stabilized conditions, not as a guarantee of robustness on live storefronts. More broadly, the generated shops are behaviorally aligned approximations rather than exact reconstructions, so some fine-grained brand-specific, visual, or platform-specific cues are intentionally not preserved.

Our validation metrics are proxies for realism: the structural analysis uses accessibility-tree depth, interaction counts, and transition-graph statistics, which capture meaningful aspects of interface and navigation complexity, but do not fully measure semantic ambiguity, visual fidelity, content quality, or all sources of user-facing behavioral variation. The structural comparison in \S\ref{sec:structural-validity} is also unpaired and aggregated over a small number of shops (seven real, three synthetic), so it characterizes central tendency rather than the full distribution; the per-shop ranges in Table~\ref{tab:shop-diversity} are the more informative view of spread, and a larger paired sample would be needed to test distributional overlap directly.

The behavioral comparison in \S\ref{sec:results-codegen} shows strong agreement in aggregate: success rates on real and twin shops track each other across models and horizons, per-task-family profiles are highly correlated, and failure-mode distributions are similar. But agreement is not identity. Matched real and twin runs of the same task produce the same outcome in only $73.2\%$ of cases, and per-task-family gaps are largest on policy lookup. Aggregate correlation therefore indicates that real and twin shops exercise the same broad skills at comparable difficulty; it does not establish that an individual task probes the same latent capability in both variants. Our scalability measurements are likewise reported over ten generated shops with specific harness and model configurations, and per-shop cost and wall-clock time will vary with the seed storefront, the coding-agent harness, and provider pricing. The parallel-rollout experiment in Appendix~\ref{sec:infrastructure-requirements} demonstrates systems feasibility rather than a completed reinforcement-learning evaluation.

\textbf{Anonymization.} Anonymization depends on the exploration agents excluding source-specific information. Because only exploration accesses the source storefront and generation sees only the specification $M$, any residual source-specific information must already appear in $M$, making it an explicit checkpoint at which a human can remove brand-, product-, or design-specific details. Multiple independent LLM judges can additionally flag potentially identifiable content for human inspection, and synthetic brand identities come from a predefined allowlist rather than unconstrained generation (Appendix~\ref{app:brand-allowlist}). Because source-brand identity is omitted, ShopGym also does not test associations with specific real-world brands, although agents must still distinguish brand and vendor information from product attributes.

Finally, the current task suite covers only part of the e-commerce agent problem. ShopGuru focuses on product discovery, browsing, filtering, information seeking, and multi-step in-store shopping journeys. It does not yet cover several important settings, including login-protected flows, checkout and payment, recommendation modules and sponsored content, customer-service interactions, cross-store comparison, post-purchase workflows, multi-session memory, multilingual settings, or safety- and policy-critical edge cases. Our tasks are also intentionally concrete, grounded in observable shop state, and automatically verifiable: they specify the required outcome while leaving the workflow (locating controls, navigating, selecting products, handling UI noise, recovering from errors, and managing the cart) to the agent. They do not yet model high-level, preference-rich intents with multiple acceptable outcomes; for example, given ``shop for $X$; you value quality,'' one agent may browse without buying while another adds an item to the cart, and both may be reasonable, making a single binary criterion ambiguous. Because ShopGuru decouples task authoring from shop generation, such intents can be supported without changing ShopArena or the generated environments, for example by making the task-generator prompt more abstract or incorporating exploration-based task generation~\cite{ramrakhya2026scaling,huang2026gta}, but they require evaluation protocols that accommodate multiple valid outcomes, clarification, and abstention. Finally, while the orchestration pipeline is domain-agnostic, its current components are tailored to e-commerce: extending ShopGym to another domain such as banking would require adapting the offline indexer used during exploration, replacing the starter website template supplied to the generation agents, and providing a suitable backend. Extending the framework to these settings is an important direction for future work.

\clearpage
\section{Broader Impact and Safeguards}
\label{sec:broader-impact}
ShopGym is intended to improve the reproducibility, controllability, and inspectability of e-commerce web-agent evaluation. By converting live storefronts into sandboxed evaluation environments, the framework reduces reliance on non-stationary live websites and enables more reliable comparison across agents.

\textbf{Potential positive impact}. ShopGym is a framework for constructing e-commerce simulation environments and grounded benchmark tasks. The simulation layer, ShopArena, provides a framework for creating diverse, stable and realistic simulation environments from live storefronts. Beyond evaluations, the environments can be used for any agent simulations, such as training of e-commerce agents using Reinforcement Learning.

\textbf{Potential negative impact}. We nevertheless identify several potential negative broader impacts. First, progress on web-agent evaluation may indirectly contribute to stronger agents that could be misused for unauthorized browsing automation, large-scale scraping, spammy shopping behavior, inventory manipulation, or deceptive commercial activity. Second, because the pipeline uses live storefronts during the exploration phase, careless use could raise privacy, copyright, fair-use, or terms-of-service concerns if raw source-store content, identifiable commercial information, or original assets were copied or redistributed. Third, generated benchmarks may not be representative of all e-commerce websites, merchant categories, languages, geographic regions, accessibility patterns, or user populations. Claims about benchmark diversity should therefore be interpreted relative to the seed storefronts and task families used to instantiate a particular ShopGym benchmark.

\textbf{Mitigations}. We mitigate these risks through several safeguards. ShopGym is designed as a sandboxed research environment rather than a tool for acting on live commercial websites. Generated shops are self-contained, locally hostable, resettable, and inspectable. Login is disabled and checkout is mocked, so the benchmark does not support real purchases, account access, or interaction with real customers. During exploration, the pipeline extracts public, user-facing structural information such as navigation affordances, catalog statistics, page organization, and policy-page structure. Downstream generation uses anonymized specifications and does not preserve brand names, product names, named individuals, contact details, or absolute URLs. Released benchmark artifacts use synthetic brand identities, products, descriptions, and imagery rather than direct copies of source-store assets.

\textbf{Synthetic brand identities}.\label{app:brand-allowlist} ShopGym intentionally omits the source storefront's brand identity and fine-grained brand-specific cues, reducing privacy, trademark, and licensing concerns. The sandbox shops are not brand-free: each receives a synthetic identity, and store and product-vendor names are drawn from a predefined allowlist screened for collisions with existing brands. Other identity attributes, including tone, tagline, and visual metadata, are synthesized. Because the allowlist is configurable, benchmark creators may replace or extend it with synthetic brands or authorized real brands; omitting source-brand information is therefore a default policy, not an architectural restriction.

We do not release raw source-store data or original commercial assets. Users who instantiate ShopGym on additional storefronts should respect applicable website terms, robots policies where relevant, copyright restrictions, and regional legal requirements. These safeguards are intended to support responsible research use while reducing the risk that the framework is used for unauthorized live-web automation or redistribution of commercial content.

\clearpage
\section{ShopArena Generation Artifact Example}
\label{sec:shoparena-generation-example}

This appendix demonstrates the sandbox shops generated by the ShopArena shop generation pipeline. These sandbox shops are built from exploration artifacts generated from ShopArena exploration tasks.

\begin{figure}[!htbp]
    \centering
    \begin{subfigure}[t]{0.32\textwidth}
      \centering
      \includegraphics[width=\linewidth]{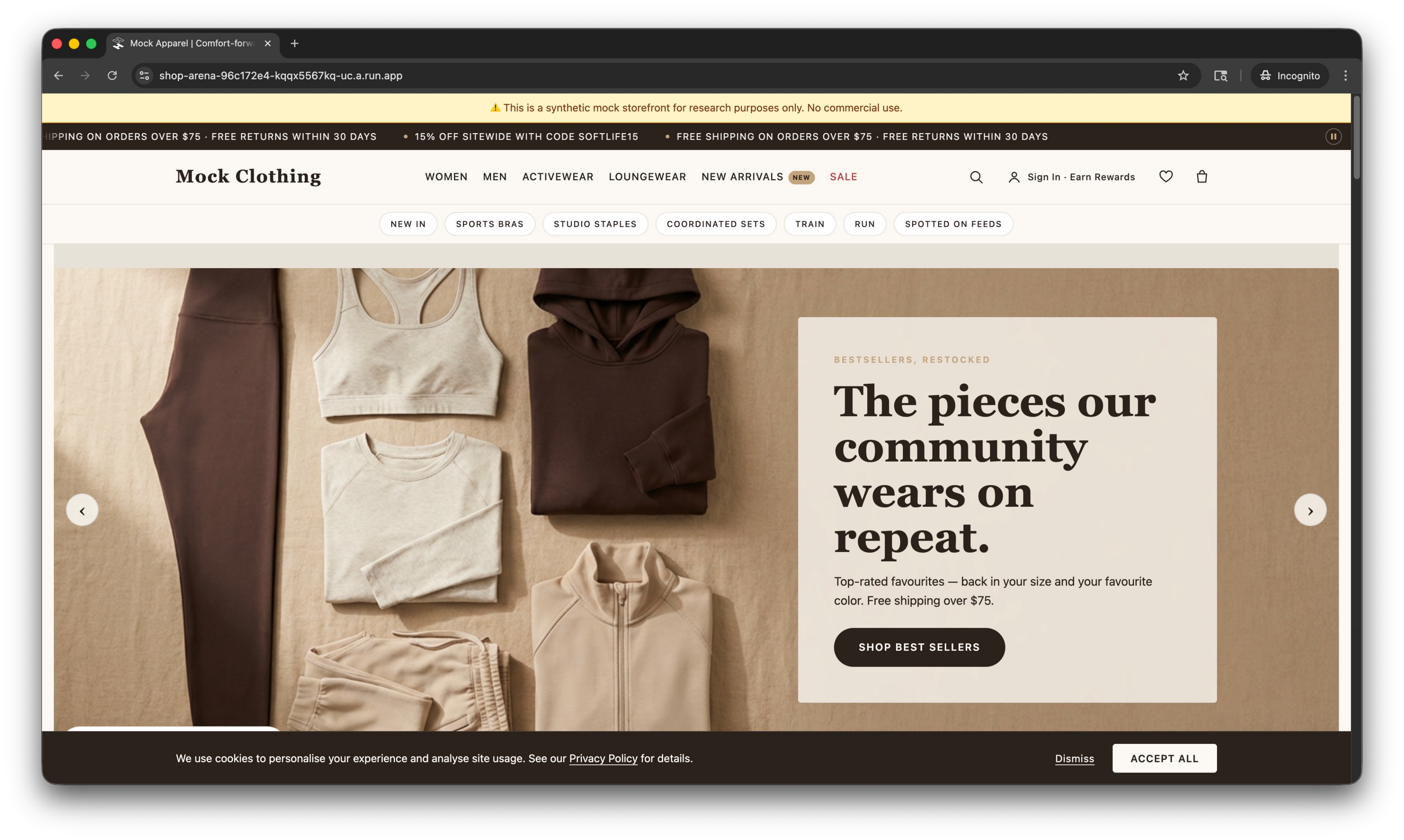}
      \caption{Mock Clothing}
      \label{fig:screenshots-mock_clothing}
    \end{subfigure}
    ~
    \begin{subfigure}[t]{0.32\textwidth}
      \centering
      \includegraphics[width=\linewidth]{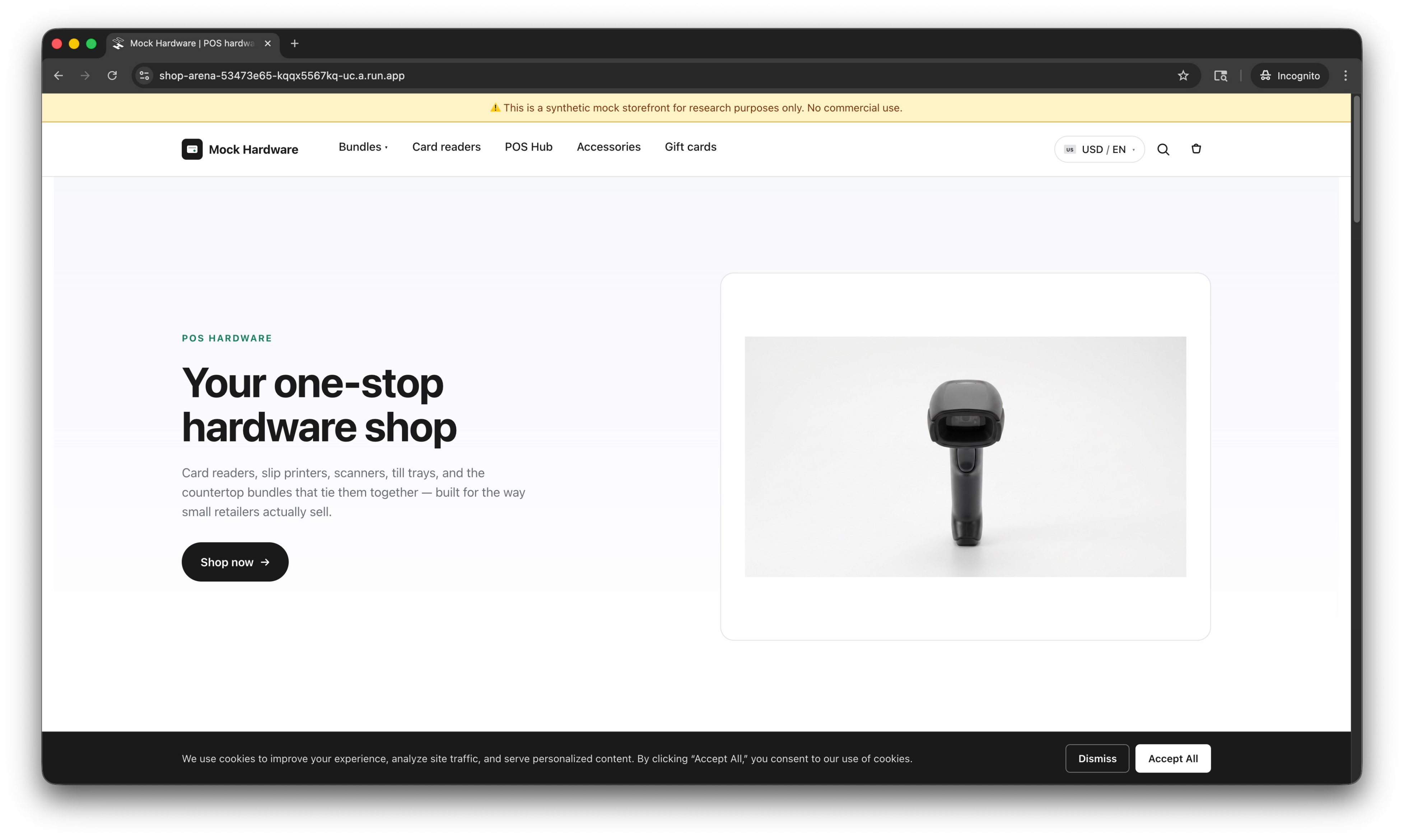}
      \caption{Mock Hardware}
      \label{fig:screenshots-mock_hardware}
    \end{subfigure}
    ~
    \begin{subfigure}[t]{0.32\textwidth}
      \centering
      \includegraphics[width=\linewidth]{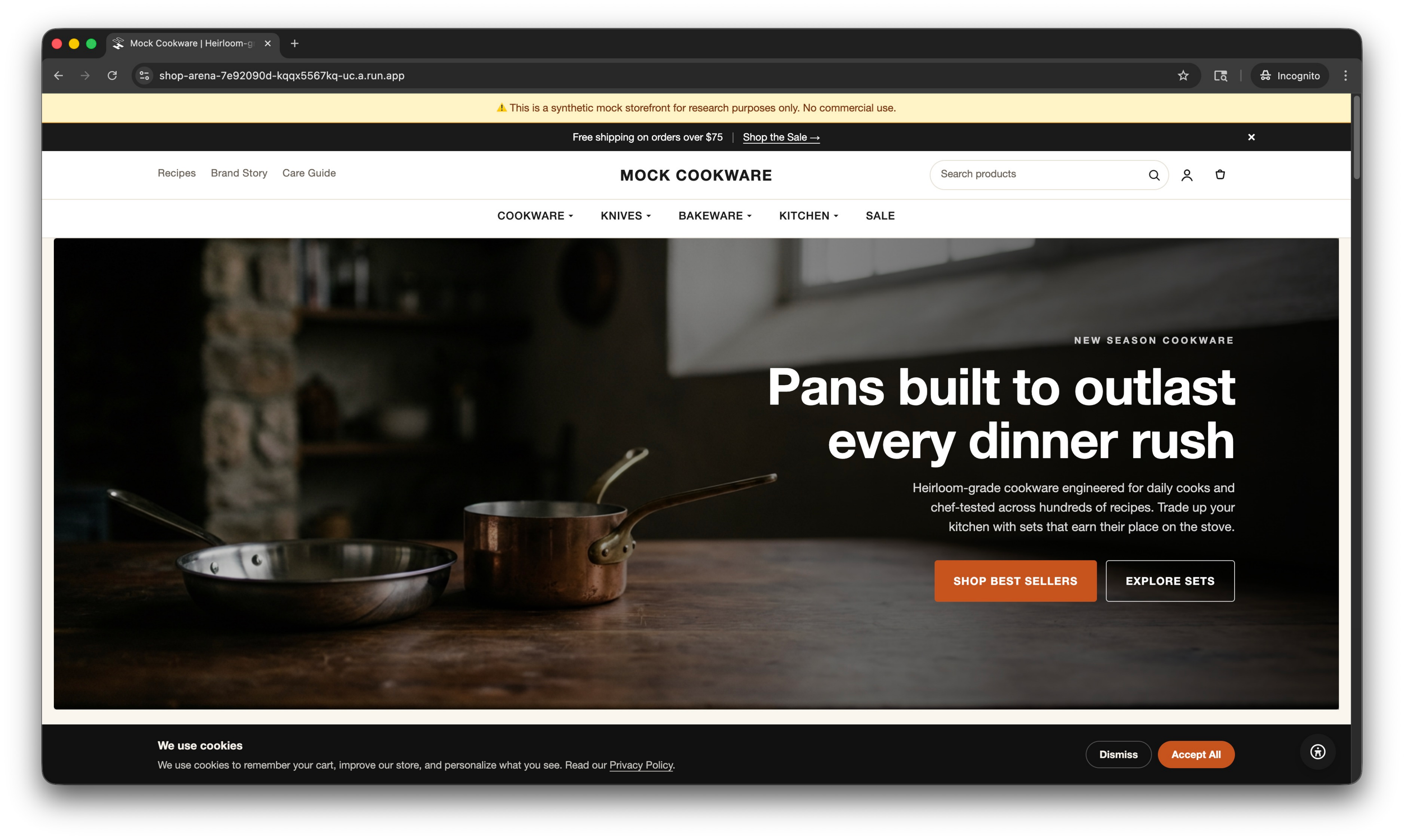}
      \caption{Mock Cookware}
      \label{fig:screenshots-mock_cookware}
    \end{subfigure}
    \caption{Screenshots of Sandbox Shop Homepages}
    \label{fig:screenshots-mock_shops}
\end{figure}

\begin{figure}[!htbp]
    \centering
    \begin{subfigure}[t]{0.32\textwidth}
      \centering
      \includegraphics[width=\linewidth]{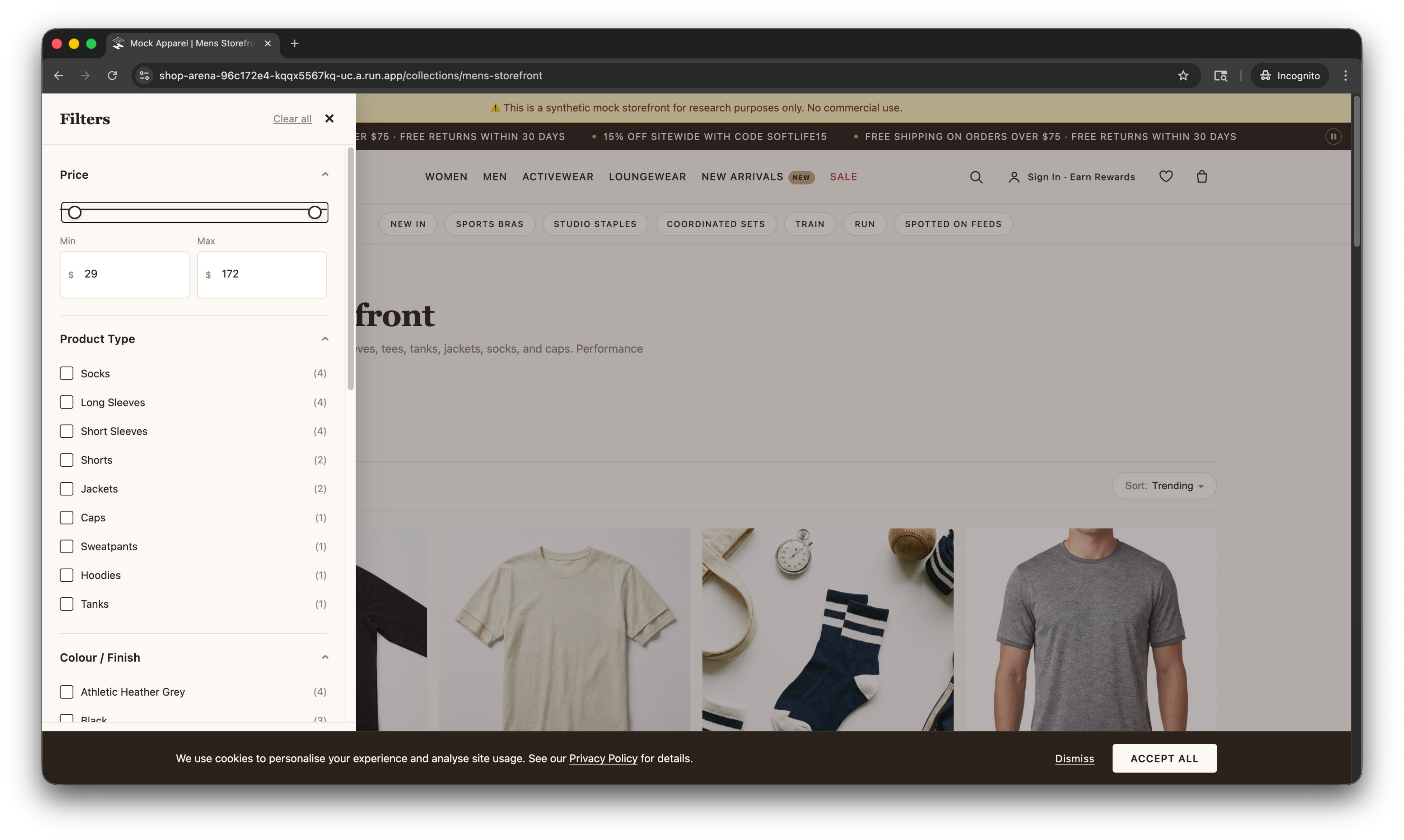}
      \caption{Collection page with filters}
      \label{fig:screenshots-mock_clothing_collection}
    \end{subfigure}
    ~
    \begin{subfigure}[t]{0.32\textwidth}
      \centering
      \includegraphics[width=\linewidth]{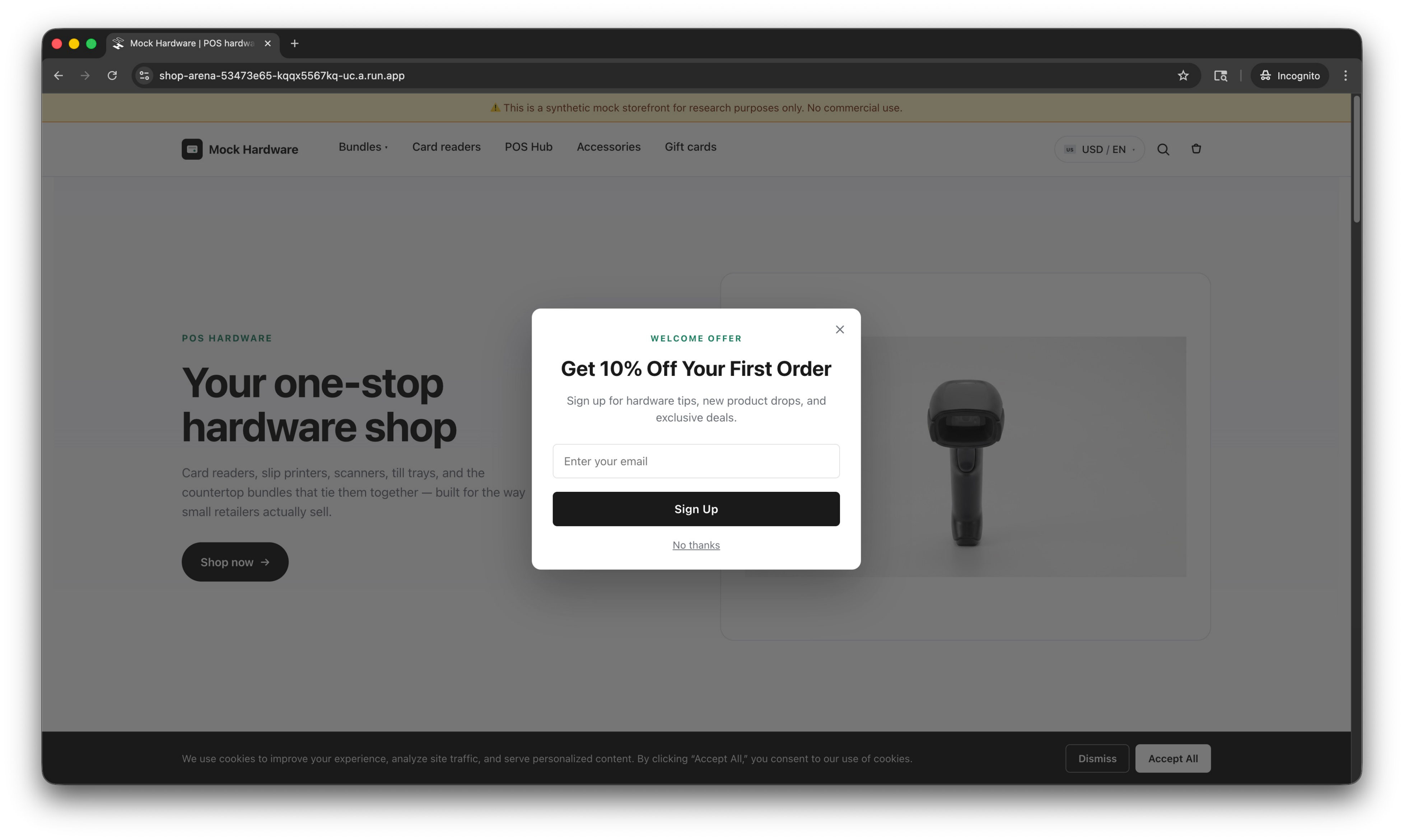}
      \caption{Homepage with promotion popup}
      \label{fig:screenshots-mock_hardware_popup}
    \end{subfigure}
    ~
    \begin{subfigure}[t]{0.32\textwidth}
      \centering
      \includegraphics[width=\linewidth]{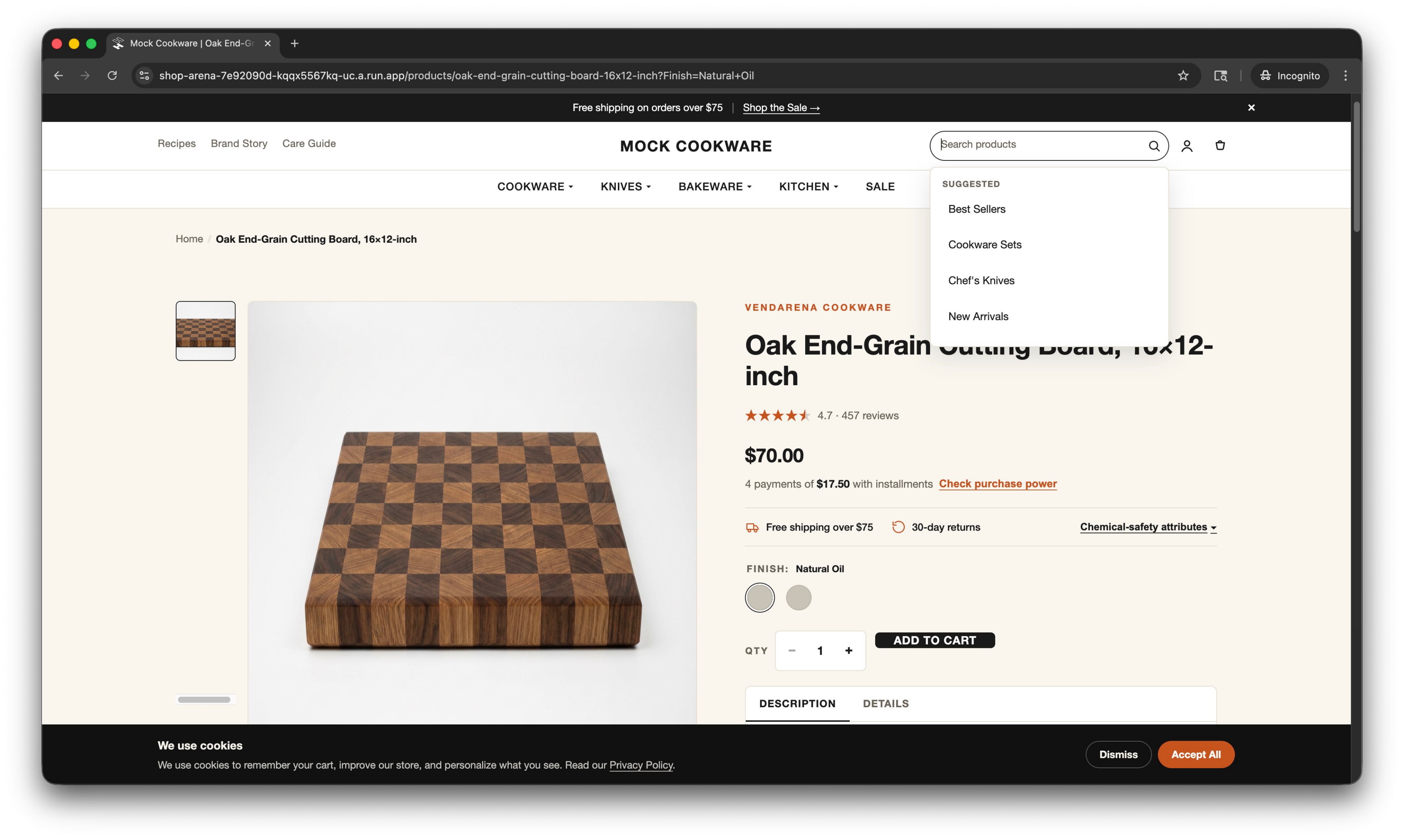}
      \caption{Product detail page}
      \label{fig:screenshots-mock_cookware_pdp}
    \end{subfigure}
    \caption{Screenshots of Sandbox Shop Details}
    \label{fig:screenshots-mock_shops_details}
\end{figure}

Figure \ref{fig:screenshots-mock_shops_details} shows more detailed interaction surfaces within the generated sandbox shops. Built from specifications, the examples include a collection page with faceted filters, a homepage with a promotional popup, and a product detail page with search suggestions and purchasing controls. These screenshots highlight that the generated shops contain not only realistic visual layouts, but also functional e-commerce elements that support browsing, filtering, promotion handling, search, and product-level decision making.

Live versions of the sandbox shops can be accessed at:
\begin{itemize}[leftmargin=*]
    \item Mock Clothing: \url{https://shop-arena-96c172e4-kqqx5567kq-uc.a.run.app/}
    \item Mock Hardware: \url{https://shop-arena-53473e65-kqqx5567kq-uc.a.run.app/}
    \item Mock Cookware: \url{https://shop-arena-7e92090d-kqqx5567kq-uc.a.run.app/}
\end{itemize}

\clearpage

\section{ShopArena Exploration Artifact Example}
\label{sec:specifications-example}

This appendix reproduces the three top-level outputs that
\texttt{shop\_arena.explore} writes for a single live storefront
exploration run. The example is produced from a premium hybrid
cookware storefront; all store, brand, designer, product, and URL
strings are anonymized at write time per the explorer's
constitution, so only structural and behavioral facts remain.

\subsection{Shop Manual (\texttt{manual.md})}
\label{app:shoparena-manual}
The manual instructs the generation pipeline to implement the cookware shop by dividing it into 7 features: {\tt Site Shell}, {\tt Homepage}, {\tt Collections \& Navigation}, {\tt Cart}, {\tt Search}, {\tt Floating UX}, {\tt Policy \& info pages}.

\begin{lstlisting}[style=shopguru-base]
# Shop Manual -- Premium Hybrid Cookware and Kitchen Accessories Shop

## Overview

This is a premium hybrid cookware and kitchen accessories storefront priced in USD. The brand voice is bold, chef-endorsed, and direct-to-consumer, leaning on culinary-professional credibility and a "buy once" durability narrative. Aggregate review counts in the 50,000+ range and a stated lifetime warranty signal a sizeable, established catalog rather than a niche boutique.

## Site shell

A persistent dismissible announcement bar sits above a two-row sticky header. Row 1 is a utility bar: a left cluster of three editorial text links plus a "Partners" dropdown button listing 2-3 named collaborator pages, a centered logo, and a right cluster of search, account, and cart icon buttons. Row 2 is the primary category navigation -- eight top-level entries rendered as inline buttons. Most entries open flyout panels rather than a single mega menu, so navigation depth tops out at two levels (top-level -> sub-link). Each flyout is a full-width two-column panel: a vertical sub-link list ending in "Shop All [Category]" on the left, plus 2-3 featured product cards on the right (image, name, price with optional compare-at, savings badge, and star rating). One top-level entry behaves as a direct link instead of opening a flyout.

A mobile off-canvas drawer mirrors the same taxonomy as accordions, with extra direct links for sale, gift cards, "Shop All", and the Partners group.

The footer occupies a `contentinfo` landmark on every page and is split into a top band and a bottom band. The top band has three named link groups (Shop, Help, Media) plus a fourth column of four social icon links opening in new tabs. The bottom band is a flex row pairing a copyright line with two legal links (terms of service, accessibility commitment) against a brand wordmark image. The footer carries no inline newsletter form, no payment icons, and no locale or currency switcher.

## Homepage

The homepage is a long, modular composition of sixteen sections in this order:

- Full-width hero with autoplay muted looping background video and a centered headline + single CTA overlay.
- "Shop by Category" chip slider -- a horizontally scrolling row of seven image-backed sub-collection chips with prev/next arrows and a "Shop All" link.
- Featured sets promo block -- text-left / product-cards-right, with a 3-card row showing image, savings badge, name, current price, and original price.
- Specialty product banner -- a reversed two-column block pairing two product cards on the left with descriptive copy on the right introducing a limited-time sub-line.
- Chef endorsement block -- autoplay video thumbnail with a play overlay opening fullscreen playback, paired with an attributed blockquote and a star + aggregate-review-count social-proof line.
- 2x2 feature card grid -- four cards covering high-heat capability, PFAS-free nonstick, dishwasher safety, and durability/warranty, each with a small attribute badge, an all-caps headline, and a one- to two-sentence descriptor.
- Reviews banner -- aggregate rating + rotating customer quote on one side, a three-icon trust strip (lifetime warranty, free shipping, 30-day returns) on the other.
- "Best Sellers" horizontal product carousel with prev/next arrows and a "Shop All" link.
- Knife category hero with two product cards on the left and a category value-prop + customer quote chip on the right.
- UGC video carousel -- six video thumbnail tiles with play overlays; tiles open a fullscreen popup player.
- 3-up specialty category banners -- three equal-width cards each combining a product image, headline, descriptor, CTA link, and a customer quote chip with star rating.
- Culinary council section introducing the brand's chef advisory group with five professional names (roles only) and a CTA to a dedicated page.
- "Recipes" carousel of recipe cards with image, title, yield, and time, linking into the recipes blog.
- "Articles" carousel of editorial blog cards using the same card chrome minus the recipe metadata.
- Founder message -- a short two-paragraph first-person note signed with a plain-text name.
- Inline newsletter section at the page bottom with an email input, hidden honeypot, and "Subscribe" button.

There is no homepage modal popup. A separate Klaviyo-driven popup is configured but did not fire during exploration.

## Collections & navigation

Collection pages render as a three-column grid on desktop. Each card shows a product image, an optional promotional overlay badge ("SALE", "New", "Bestseller"), product title, star rating with review count, and price (strikethrough original price plus a percentage-off badge when discounted). The collection title displays a live product count in parentheses.

Above the grid sits a horizontal filter bar with two collapsible controls. The "Filter" panel exposes two checkbox facets -- "Available" and "On Sale" -- that update URL parameters immediately, render server-side, and combine cleanly. The "Sort by" dropdown offers eight options: featured, best selling, alphabetically A-Z, alphabetically Z-A, price ascending, price descending, date new-to-old, and date old-to-new. Sorting updates the `sort_by` URL parameter without a full reload.

Pagination uses a "Load more" button. The initial render shows 24 cards; clicking the button appends the rest of the collection in a single batch and the button then disappears. There is no infinite scroll, no numbered pagination, and no `page=` URL parameter.

## Product page

The product page leads with a horizontal image carousel that has prev/next arrow controls and an optional left-side thumbnail strip for direct slide selection. Below the title sits the price block: prominent current price, optional strikethrough original price, optional percentage savings badge, and a buy-now-pay-later financing line that opens a modal. A trust badge row beneath the price summarizes free shipping, a 30-day return guarantee, and (where applicable) a lifetime warranty, plus an expandable claim button listing chemical-safety attributes for nonstick items.

Variant selectors render as radio buttons in a single horizontal row. Apparel exposes a size axis (XS-XXL) with a "Size Guide" link above the group. Single-variant products skip the selector entirely. A quantity stepper (decrement, numeric display, increment) sits immediately adjacent to the "Add to Cart" CTA.

Descriptions vary by category. Simpler products render an inline prose paragraph plus a bulleted feature list. Complex products use an accordion-tab layout (e.g. "Description" + "Hybrid Technology" tabs) followed by a horizontal seven-icon attribute strip (warranty, utensil safety, nonstick, oven-safe, dishwasher-safe, induction-ready, stay-cool handle). Some PDPs include a "Gift With Purchase" callout above the quantity stepper showing a bundled free item.

Reviews use an Okendo-style widget. Products without reviews show a "Be the first to leave a review" prompt and a "Write a review" CTA. Products with reviews show an aggregate score, a clickable star-distribution breakdown, an auto-generated AI summary block, search and sort/filter dropdowns (latest, oldest, highest, lowest, most helpful; rating 1-5 or all), and individual review cards with reviewer name, star rating, relative date, headline, body, and helpful-vote buttons.

Recommendation rails differ by template. Some PDPs render a "Customers also love" rail of product cards with quick-add buttons and inline color-swatch selectors. Others render two stacked rails -- a "Frequently Bought Together" trio with inline "Add to Cart" buttons, and a "Recently Viewed" carousel with prev/next arrows. No customization or personalization fields (engraving, monogram, gift note) are exposed on the PDP.

## Cart

The cart opens as a right-side slide-out drawer (ARIA `dialog` labeled "Your Cart") triggered by the header cart icon. The icon shows a numeric item-count badge once the cart has any items.

Empty state shows the "Your Cart" heading with a `0` badge, a short empty-cart sentence, a CTA link into a best-sellers collection, a `$0.00` subtotal line, and the same trust-badge trio (lifetime warranty, free shipping, 30-day returns) repeated near the footer.

Filled state replaces the empty copy with a free-shipping qualifier banner once the threshold is met. Each line item shows a thumbnail, name (both linked to the PDP), the price block (current price plus strikethrough compare-at when discounted), a "Remove Item" affordance, and a three-button quantity stepper. Quantity changes apply via Ajax and immediately recompute three summary rows: products at list price, total savings, and subtotal. A full-width "Checkout" button sits below, with a footnote asterisk for free-shipping eligibility and the trust badges repeated.

The drawer carries a "Frequently Bought Together" upsell rail (three product cards with inline "Add to Cart") in both states; carted items are swapped out of the rail for fresh suggestions. A "Recently Viewed" rail with prev/next arrows appears beneath. There is no inline promo-code field -- discount codes are collected on the native checkout -- and there is no inline shipping estimate calculator; free-shipping eligibility is conveyed only via the qualifier banner and footnote.

## Search

Search is triggered by a magnifying-glass icon button in the header utility row. Clicking it expands an inline dropdown directly below the icon -- the page is not dimmed and navigation remains visible.

Predictive search renders a static "Suggested" panel inside the dropdown: typically four links combining one collection shortcut and three pre-set search-query links. The list is fixed and does not change with keystrokes. The dropdown itself does not render product thumbnails, prices, or product cards -- only suggested queries and collection shortcuts.

A `/search/suggest.json` predictive endpoint is active in the background and returns product-shaped JSON (title, handle, price, compare-at range, availability, URL, vendor, featured image, tags, type) but its results are not surfaced in the header dropdown. Submitting the form navigates to a full-page `/search?q=<query>` results listing. A second search input appears in the mobile off-canvas drawer with the same form action and no predictive panel.

## Floating UX

A third-party CMP renders a full-width cookie consent banner at the bottom of the viewport on first visit. The banner offers a privacy-policy link, "Allow All", "Reject All", and "Confirm My Choices" buttons. Granular preference controls cover four consent categories (analytics, marketing, preferences, sale of data). Choice is stored in localStorage and integrates with Google Consent Mode V2 for downstream tag gating.

A Klaviyo-powered newsletter popup is configured (loaded via async script). It is time- or session-gated and did not appear during cold-session exploration. The inline newsletter section in the page footer area remains the always-on capture surface.

No chat widget, no age gate, and no homepage modal popup were observed.

## Policy & info pages

Seven informational pages are exposed. Four are native policies under `/policies/<slug>` -- shipping, refund, privacy, and terms of service -- all sharing a minimal single-column template: H1 title, then rich text paragraphs, H2 sub-sections, and bullet lists, with no images or sidebars.

Three custom pages live under `/pages/<slug>`. The contact page uses a hero split layout with a headline, a one- to two-sentence body, and a CTA button deep-linking into an external customer-support portal (no inline form), followed by a three-card informational grid covering warranty, product registration, and 24/7 support. The FAQ page uses a custom anchored-section template: an H1, a horizontal in-page navigation bar with four anchors (Products and Usage, Orders, Contact, Warranty), and four H2 sections of accordion-style question/answer items. The about page is a custom editorial layout with a celebrity-chef hero quote, a brand-story section pairing copy with imagery, a five-row numbered feature list, a three-link "Shop" CTA row, and an inline email subscribe form with a hidden honeypot.
\end{lstlisting}

\subsection{Capabilities (\texttt{capabilities.json})}
\label{app:shoparena-caps}

\begin{lstlisting}[style=shopjson]
{
  "version": "0.1",
  "shop": {
    "descriptor": "Premium hybrid cookware and kitchen accessories shop",
    "category": "Cookware / Kitchen",
    "currency": "USD",
    "tone": [
      "premium",
      "bold",
      "chef-endorsed",
      "direct-to-consumer"
    ]
  },
  "site_shell": {
    "has_announcement_bar": true,
    "header_style": "two-row sticky",
    "has_mega_menu": false,
    "nav_depth": 2,
    "footer_groups": 3
  },
  "homepage": {
    "section_types": [
      "hero_video",
      "category_chip_slider",
      "featured_product_grid",
      "specialty_product_banner",
      "video_endorsement",
      "feature_cards_grid",
      "reviews_banner",
      "product_carousel",
      "product_hero_quote",
      "ugc_video_carousel",
      "category_banners_3up",
      "culinary_council",
      "recipe_carousel",
      "article_carousel",
      "founder_message",
      "newsletter_inline"
    ],
    "section_count": 16,
    "has_popup_modal": false
  },
  "collection": {
    "layout": "grid",
    "columns_desktop": 3,
    "filters": [
      "availability",
      "on_sale"
    ],
    "sort": [
      "featured",
      "best_selling",
      "alphabetically_az",
      "alphabetically_za",
      "price_asc",
      "price_desc",
      "date_new",
      "date_old"
    ],
    "pagination": "load_more"
  },
  "product": {
    "gallery_style": "horizontal_carousel_with_prev_next",
    "variant_selectors": [
      "radio_buttons"
    ],
    "has_quantity_selector": true,
    "description_layout": "accordion_tabs",
    "has_reviews": true,
    "has_recommendations": true,
    "has_personalization": false
  },
  "cart": {
    "type": "drawer",
    "has_promo_input": false,
    "has_upsells": true,
    "has_shipping_estimate": false
  },
  "search": {
    "trigger": "icon_button",
    "has_predictive": true,
    "predictive_types": [
      "suggested_queries",
      "collection_shortcuts"
    ],
    "results_layout": "fullpage"
  },
  "intl": {
    "has_locale_switcher": null,
    "has_currency_switcher": null
  },
  "floating": {
    "has_chat_widget": false,
    "has_age_gate": false,
    "has_cookie_banner": true,
    "has_newsletter_popup": true
  },
  "info_pages_present": [
    "shipping-policy",
    "refund-policy",
    "privacy-policy",
    "terms-of-service",
    "contact",
    "faq",
    "about-us"
  ]
}
\end{lstlisting}

\subsection{Statistics (\texttt{stats.json})}
\label{app:shoparena-stats}

\begin{lstlisting}[style=shopjson]
{
  "products_total": 164,
  "collections_total": 50,
  "products_per_collection": {
    "avg": 15.42,
    "median": 8.0,
    "max": 179
  },
  "price": {
    "min": 0.98,
    "max": 5499.0,
    "median": 59.99,
    "currency": "USD"
  },
  "products_with_variants_pct": 0.07317073170731707,
  "variant_axes_observed": [
    "title",
    "size"
  ],
  "navigation_depth_max": 2,
  "homepage_section_count": 16,
  "info_pages_count": 7,
  "feature_count": 48
}
\end{lstlisting}

\clearpage

\section{ShopArena Exploration Plan Example}
\label{app:shoparena-plan}

This appendix reproduces the \texttt{plan.md} document emitted by
the planner iteration of \texttt{shop\_arena.explore} for the same
exploration run shown in Appendix~\ref{sec:specifications-example}. The
plan enumerates the per-task work items the executor loop then
consumes one task at a time; checkbox state (\texttt{[x]} /
\texttt{[!]}) is updated by the executor at the end of each
iteration.

\subsection{Plan (\texttt{plan.md})}
\label{fig:shoparena-plan-md}

\begin{lstlisting}[style=shopguru-base]
# Plan -- Premium Hybrid Cookware Shop

A direct-to-consumer shop specializing in premium hybrid cookware, knives, kitchen tools, seasonings, and branded apparel, endorsed by a celebrity chef. Sells via USD pricing with US-focused shipping. The shop uses a two-row sticky header, flyout dropdown navigation, and a cart drawer architecture.

## Tasks

- [x] homepage_sections -- navigate to `/`; capture hero section, each below-hero section (category grid, endorsement block, features section, best-sellers carousel, social-proof/testimonials, specialty-product banners, recipe carousel), note any newsletter popup behavior; full-page screenshot + per-section evidence [priority: 9]
- [x] cart_drawer -- add 1 product to cart; capture drawer in empty state, filled state, qty-change state; check for promo-code input, upsell rail, and shipping-estimate widget [priority: 8]
- [x] header_navigation -- snapshot main nav in closed state, then trigger each top-level flyout button (8 categories); capture one open flyout; record nav depth and sub-link count per category [priority: 7]
- [x] search_predictive -- open search modal; type 2 short prefixes ("s" and "pan"); capture suggestion panel showing products; capture one network request via network.jsonl [priority: 7]
- [x] collection_filters -- visit the largest cookware collection; exercise "Available" and "On Sale" filter checkboxes; test all sort options; scroll to bottom to confirm pagination behavior; screenshot grid and record column count [priority: 6]
- [x] product_variants -- visit 2 PDPs: one apparel product (has size radio buttons) and one cookware set (single-variant); capture gallery carousel states (prev/next), variant selector states, qty stepper, description layout, reviews widget (Okendo), and "Customers also love" recommendations rail [priority: 5]
- [x] footer -- screenshot full footer; record link groups, social icons, and bottom-bar copyright/legal links [priority: 4]
- [x] info_pages -- visit `/policies/shipping-policy`, `/policies/refund-policy`, `/policies/privacy-policy`, `/policies/terms-of-service`, `/pages/contact`, `/pages/faq`, `/pages/about-us`; one screenshot each; note page layout type (policy template vs custom page) [priority: 3]
- [x] floating_widgets -- wait for and capture Klaviyo newsletter popup; inspect Crisp chat widget launcher; check OneTrust cookie banner presence; one screenshot per widget if visible [priority: 1]

## Omitted Areas

- mega_menu -- nav uses flyout dropdown panels per category (single-column sub-link lists), not a full-grid mega menu layout; observed in homepage snapshot
- intl_switchers -- only `en_US` locale present; no locale or currency switcher visible in header, footer, or homepage snapshot; `index.html` locale metadata is `en_US` only
- age_gate -- not observed on `/` homepage snapshot or in `index.html` HTML structure
\end{lstlisting}

\clearpage
\section{ShopGuru Skill Catalog}
\label{app:shopguru-skills}

This appendix accompanies \S\ref{sec:shopguru-skills} and summarizes the seven skill categories generated by ShopGuru.

\begin{table}[h!]
\centering
\caption{ShopGuru's seven skill categories grouped by horizon.
Short-horizon tasks are produced by deterministic, pure-function
generators over ShopArena's extracted JSON; long-horizon tasks are
LLM-authored from few-shot examples reflecting human-shopper behaviors
and reconciled against the same extracted data via a validator-driven
polish loop (\S\ref{sec:shopguru-e2e}).}
\label{tab:shopguru-skills}
\small
\begin{tabular}{p{0.18\linewidth} p{0.16\linewidth} p{0.22\linewidth} p{0.32\linewidth}}
\toprule
\textbf{Group} & \textbf{Bundle} & \textbf{Source data} & \textbf{Skill primitives} \\
\midrule
Discovery & short-horizon & \texttt{products.json} & exact title search, substitute search \\
\addlinespace[2pt]
Filter-Selection & short-horizon & \texttt{collections.json} \,+\, per-collection options & collection browse, collection filter \\
\addlinespace[2pt]
Information-Seeking & short-horizon & \texttt{pages.json} (+ default policy fallback) & shipping policy lookup, returns policy lookup \\
\addlinespace[2pt]
E2E shopping journey & long-horizon & all of the above + few-shot human-trace examples & LLM-authored multi-step intents \\
\bottomrule
\end{tabular}
\end{table}

\clearpage

\section{Examples of ShopGuru tasks}

\begin{lstlisting}[style=shopjson,
caption={Example collection-filter task emitted by the deterministic
generator. \texttt{url\_contains} is a soft URL hint to the LLM judge,
not a hard pass/fail gate.},
label={lst:filter-task}]
{
    "id": "mock_clothing-filter-1",
    "type": "shopping",
    "intent": "Navigate to the \"Frost Season Collection\" collection on this store. Find and use the Color filter (e.g. Black) to select an option. If products are shown after filtering, select any variant of a product and add it to cart. It is ok if no products match the filter; using filter correctly means the navigation task is completed successfully.",
    "success_criteria": {
      "url_contains": "/collections/frost-season-collection",
      "type": "navigation"
    },
}
\end{lstlisting}

\begin{lstlisting}[style=shopjson,
caption={Example page-navigation task to find specific policy on the shop, also emitted by the deterministic
generator.},
label={lst:page-navigation-task}]
{
    "id": "mock_hardware-returns-1",
    "type": "navigation",
    "intent": "Find the returns and refund policy page on this store. Look for return policy information in the footer, menu, or any navigation links. Read the refund policy details, then leave the site.",
    "success_criteria": {
      "url_contains": "/policies/refund-policy",
      "type": "page_navigation"
    },
    "url_contains_alt": [
      "/pages/return-policy"
    ],
}
\end{lstlisting}

\begin{lstlisting}[style=shopjson,
caption={Example LLM-authored long-horizon journey that chains a
collection navigation, a Color filter, a price sort option, and a
final cart action.},
label={lst:e2e-task}]
{
    "id": "mock_cookware-e2e-v1-3",
    "type": "shopping",
    "intent": "Shopping for a new chef's knife on a budget. Navigate to the Knives collection. Apply a filter for Color if available, or sort the collection by Price: Low to High. Browse the filtered or sorted results and pick any Chef's Knife. Open its page, choose whichever Color variant you like best (e.g., Black Handle or Natural Wood Handle), and add it to cart. Once the knife is in your cart, end the session. Do not click any Checkout button.",
    "success_criteria": {
      "url_contains": "/collections/knives",
      "type": "cart_after_filter_or_sort"
    }
}
\end{lstlisting}

\begin{lstlisting}[style=shopjson,
caption={Another example LLM-authored long-horizon task consisting of a shopping journey after a page navigation detour.},
label={lst:e2e-task-2}]
{
   "id": "mock_clothing-e2e-v1-1",
    "type": "shopping",
    "intent": "First-time visitor exploring the brand's ethos. Open the About Us page from the footer or main menu to read up on Mock Clothing's lifestyle positioning. Next, navigate to the Fresh Drops collection to see what's newly available. Pick any accessory (like a scrunchie or cap) that catches your eye, open its product page, select any available Color and Size variants, and add it to your cart. Once the item is in your cart, end the session. Do not click any Checkout button.",
    "success_criteria": {
      "url_contains": "/pages/about",
      "type": "cart_after_about_page_detour"
    },
}
\end{lstlisting}

\clearpage

\section{ShopGuru Polish Loop}
\label{app:shopguru-polish-loop}

This appendix accompanies \S\ref{sec:shopguru-e2e} and illustrates the LLM-authored E2E polish loop.

\begin{figure}[h!]
    \centering
    \resizebox{\linewidth}{!}{%
    \begin{tikzpicture}[
        >=Latex,
        every node/.style={font=\footnotesize, align=center, inner sep=3pt},
        box/.style={draw, rounded corners=1.5pt, minimum height=8mm},
        ctx/.style={box, fill=blue!8, minimum width=42mm, align=left},
        llm/.style={box, fill=orange!15, minimum width=32mm},
        val/.style={box, fill=red!10, minimum width=32mm},
        dec/.style={diamond, draw, fill=yellow!20, aspect=2, inner sep=1pt, font=\scriptsize, align=center},
        outbox/.style={box, fill=green!12, minimum width=24mm},
        polishbox/.style={box, fill=purple!8, minimum width=42mm, align=left},
        arr/.style={->, thick, shorten >=1pt, shorten <=1pt},
        loopback/.style={->, dashed, thick, shorten >=1pt, shorten <=1pt},
    ]
    \node[ctx] (ctx) {Shop context\\\scriptsize(catalog, collections, pages)\\$+$ system prompt $+$ 8 few-shot examples};

    \node[llm, right=10mm of ctx] (llm0) {LLM (round $k$)\\\scriptsize e.g.\ GPT-5, Gemini 3 Pro};

    \node[val, right=8mm of llm0] (val) {Validator\\\scriptsize(7 rules)};

    \node[dec, right=8mm of val, minimum width=14mm, minimum height=8mm] (dec) {issues?};

    \node[outbox, right=8mm of dec] (final) {final\\tasks};

    \node[polishbox, below=10mm of val] (polish) {Polish prompt:\\\scriptsize$\bullet$ flagged tasks (preserve IDs)\\\scriptsize$\bullet$ specific validator issues\\\scriptsize$\bullet$ shop context};

    \draw[arr] (ctx.east) -- (llm0.west);
    \draw[arr] (llm0.east) -- (val.west);
    \draw[arr] (val.east) -- (dec.west);
    \draw[arr] (dec.east) -- (final.west) node[midway, above, font=\scriptsize] {none};

    \draw[arr] (dec.south) |- (polish.east) node[pos=0.25, right, font=\scriptsize] {flagged};
    \draw[loopback] (polish.west) -| (llm0.south) node[pos=0.75, left, font=\scriptsize, align=right] {regenerate\\flagged only};

    \node[font=\scriptsize, below=2mm of polish.south, anchor=north] {at most 2 rounds, then emit};
    \end{tikzpicture}%
    }
    \caption{LLM-authored E2E polish loop. The same validator that
    audits deterministic generators (\S\ref{sec:shopguru-validate})
    drives an iterative refinement of LLM-authored tasks. The round-$k$
    output is audited; if any tasks are flagged, the polish prompt
    bundles \emph{only the flagged tasks} with their specific
    validator issues and the shop context, the LLM regenerates only
    those tasks while preserving their IDs, and the merged result is
    re-audited. The loop terminates after at most two rounds; tasks
    that still fail validation after two rounds are surfaced as a
    release blocker rather than silently shipped.}
    \label{fig:polish-loop}
\end{figure}
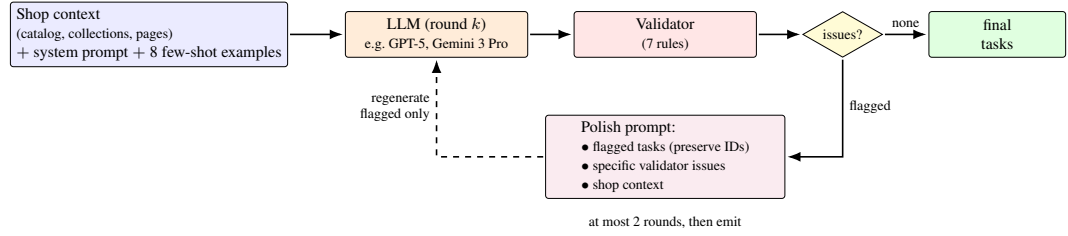

\clearpage

\section{Verifiers for ShopGuru pipeline}
\label{app:shopguru-verifiers}

This appendix accompanies \S\ref{sec:shopguru-validate} and lists
the rule set used by ShopGuru's post-generation validator. The
validator runs as the final gate on every emitted benchmark file:
\texttt{error}-severity findings halt the build with a non-zero
exit code, while \texttt{warning}-severity findings are logged
and either fed back to the LLM-authored journey generator as
polish-loop signals (rules marked $^{\dagger}$ in
Table~\ref{tab:shopguru-rules}) or surfaced as advisory hints
for the human curator. See \S\ref{sec:shopguru-validate} for the
narrative description and \S\ref{sec:shopguru-e2e} for the
polish-loop integration.

\begin{table}[h!]
\centering
\caption{ShopGuru's seven core validator rules. The validator returns
a list of \texttt{Issue} records rather than raising, so callers can
decide whether to error or just warn. Errors halt the build with a
non-zero exit code. Rules marked with $^{\dagger}$ are additionally
fed back to the LLM-authored journey generator as polish-loop signals
(\S\ref{sec:shopguru-e2e}); the unmarked warning is advisory.}
\label{tab:shopguru-rules}
\small
\begin{tabular}{l l p{0.55\linewidth}}
\toprule
\textbf{Rule} & \textbf{Severity} & \textbf{Catches} \\
\midrule
\texttt{unknown-collection}$^{\dagger}$ & error & \texttt{url\_contains} references a collection handle absent from \texttt{collections.json}. \\
\texttt{unknown-product}$^{\dagger}$ & error & Same, for \texttt{/products/$\langle$handle$\rangle$}. \\
\texttt{infeasible-filter}$^{\dagger}$ & error & Filter $(\textit{dim}, \textit{value})$ is not realized by any product in the targeted collection. \\
\texttt{intent-answer-leak}$^{\dagger}$ & error & A \texttt{success\_criteria.response\_contains} value appears verbatim in the intent --- an agent could pass by echoing the prompt without visiting the page. \\
\midrule
\texttt{option-mismatch}$^{\dagger}$ & warning & The intent asks the agent to ``select a $\langle$Option$\rangle$ variant'' for a product that has no such option (e.g.\ ``select Color'' on a gift card whose only option is denomination). \\
\texttt{product-not-in-collection}$^{\dagger}$ & warning & The intent pairs a product with a collection that does not contain it. \\
\texttt{unknown-page} & warning & Same as \texttt{unknown-collection} but for \texttt{/pages/$\langle$handle$\rangle$} (softer because some storefronts surface policies via \texttt{/policies/*} fallthrough routes). \\
\bottomrule
\end{tabular}
\end{table}

\clearpage
\section{ShopArena Pipeline Invocation Map}
\label{app:pipeline-invocations}

This appendix accompanies \S\ref{sec:shop-arena} and distinguishes
deterministic operations, direct LLM calls, and coding-agent sessions in
the ShopArena pipeline. We use the following notation: [D] denotes a
deterministic operation with no LLM; [L], one direct prompt--response LLM
call; and [A], one coding-agent session (Claude Code or Pi) that may
contain multiple LLM turns and Playwright or coding-tool calls. The
complete environment-construction pipeline for one source storefront is
\begin{center}
\footnotesize
\begin{tabular}{@{}r@{\hspace{0.6em}}l@{}}
\textit{Exploration:} & prefetch [D] $\rightarrow$ exploration planner [A] $\rightarrow$ exploration executors [A\,$\times T$] $\rightarrow$ manual synthesis [L] \\[2pt]
\textit{Generation:} & data synthesis [L] $\rightarrow$ build planner [A] $\rightarrow$ build/verify loop [A/L/D] $\rightarrow$ final evaluation [A/L/D],
\end{tabular}
\end{center}
where $T$ is the number of exploration tasks produced by the planner.
Table~\ref{tab:pipeline-invocations} gives the detailed invocation map.
We do not count a coding-agent session as a single LLM call, because its
number of model turns depends on the browsing or coding trajectory.

\begin{table}[h!]
\centering
\caption{Invocation map of the ShopArena pipeline for one source
storefront. $T$ is the number of planned exploration tasks, $C$ the set
of collections, $n_c$ the number of products in collection $c$, and $B$
the number of products per product-detail chunk.}
\label{tab:pipeline-invocations}
\scriptsize
\setlength{\tabcolsep}{3pt}
\begin{tabular}{p{0.15\linewidth} p{0.2\linewidth} p{0.31\linewidth} p{0.26\linewidth}}
\toprule
\textbf{Stage} & \textbf{LLM-dependent operation} & \textbf{Invocation count} & \textbf{Output} \\
\midrule
Prefetch & None [D] & 0 & Indexed HTML pages and product/collection data \\
\addlinespace[2pt]
Exploration planning & Playwright-enabled coding agent [A] & 1 session & Plan of page- and feature-specific exploration tasks \\
\addlinespace[2pt]
Targeted exploration & Playwright-enabled coding agent [A] & $T$ sessions, one per task & One evidence-backed prose fragment and one structured capability fragment per task \\
\addlinespace[2pt]
Shop Manual synthesis & Direct LLM completion [L] & 1 call; capability fragments are merged separately in Python [D] & A single anonymized Shop Manual \\
\addlinespace[2pt]
Identity and catalog synthesis & Direct structured-output completions [L] & 7 calls: identity/tone, store metadata, informational pages, policies, collections, product skeletons, navigation & Schema-validated JSON artifacts \\
\addlinespace[2pt]
Product-detail synthesis & Direct structured-output completion [L] & $\sum_{c \in C} \lceil n_c / B \rceil$ calls, one per chunk of $B$ products in each collection; at most one retry per chunk & Descriptions, variants, options, tags, and vendors \\
\addlinespace[2pt]
Alt-text synthesis & Direct structured-output completion [L] & $|C|$ calls, one per collection; at most one retry per collection & Product-image alt text \\
\addlinespace[2pt]
Image generation & Optional image-model call [L] & One call per generated image & Product images \\
\addlinespace[2pt]
Data validation & None [D] & 0 & Schema checks and a live backend API test \\
\addlinespace[2pt]
Build planning & Coding agent [A] & 1 session & Ordered storefront implementation plan \\
\addlinespace[2pt]
Storefront implementation & Coding agent [A] & Up to 30 sessions, one per loop iteration; a retry starts a new session & Source-code changes for the next unfinished plan task \\
\addlinespace[2pt]
Source verification & Direct LLM judges [L] & Up to one quality-judge call and one cross-task-consistency call after each applicable iteration & Structured verifier feedback \\
\addlinespace[2pt]
Visual verification & Playwright-enabled coding agent [A] & One session for the page bucket affected by the task; the whole-site visual-fix task uses one session per page bucket & Screenshots and a structured visual verdict \\
\addlinespace[2pt]
Rule verification & None [D] & 0 & TypeScript/build, route, data-use, and navigation-coverage results \\
\addlinespace[2pt]
Final evaluation & Playwright-enabled coding agent [A] plus a direct LLM judge [L] & One agent session per page bucket plus one judge call; smoke checks are deterministic & Advisory visual and operational report \\
\bottomrule
\end{tabular}
\end{table}

\clearpage

\section{Generation Failure Modes}
\label{app:generation-failures}

This appendix expands \S\ref{sec:scalability}. ShopArena is designed to
catch generation failures before they propagate, at two levels. Within a
run, the execution--verification loop (\S\ref{sec:shop-gen}) detects
functional and visual defects and returns structured corrections to the
execution agent. Across runs, recurring failures identified during human
review become reusable verifier checks and generation guidance, which is
what drives the decline in human inspection time per shop
(Table~\ref{tab:human-effort}).

We distinguish website-level defects, which the loop repairs, from
pipeline failures. We generate seven additional shops in seven
concurrent jobs. All seven complete the pipeline and produce buildable
websites, requiring $17$--$26$ execution iterations (mean $21.7$); none
reaches the 30-iteration budget, which therefore provides headroom beyond
the maximum observed requirement while bounding cost.
Table~\ref{tab:generation-failures} reports the errors caught by each
verifier and repaired by the execution agent in subsequent iterations.
Most retries address visual or quality defects ($88$ events), followed
by functional defects ($26$). No compile or build failures occur;
three runs encounter a plan-parser edge case after retry exhaustion.

\begin{table}[h!]
\centering
\caption{Errors caught by the verifiers and repaired in subsequent
iterations across the seven additional shops.}
\label{tab:generation-failures}
\small
\begin{tabular}{l l r}
\toprule
\textbf{Error class} & \textbf{Common example} & \textbf{Retry events} \\
\midrule
Functional defects (rule-based) & Broken routes or internal links & 13 \\
 & Missing navigation coverage & 5 \\
 & Incorrect cart behavior & 8 \\
\midrule
Visual and quality defects (LLM judge) & Layout, visual, or responsive differences & 55 \\
 & Page-level quality issues & 18 \\
 & Cross-page inconsistency & 15 \\
\midrule
Build failures (rule-based) & TypeScript or production-build failures & 0 \\
Pipeline failures (rule-based) & Plan-parser edge case after retry exhaustion & 3 runs \\
\bottomrule
\end{tabular}
\end{table}

\begin{table}[h!]
\centering
\caption{Active human inspection and correction time per shop across
generation batches.}
\label{tab:human-effort}
\small
\begin{tabular}{lcc}
\toprule
\textbf{Generation batch} & \textbf{Shops} & \textbf{Intervention per shop} \\
\midrule
Initial development & 3 & 60 min \\
Shops in \S\ref{sec:sandbox-shop-validation} & 3 & 30 min \\
Seven additional shops & 7 & ${\sim}10$ min \\
\bottomrule
\end{tabular}
\end{table}

\clearpage

\section{Stability Across Generation Runs}
\label{app:generation-stability}

Because generation is stochastic, we generate three additional versions
of one furniture shop from the same specification and evaluate their
structural and quality consistency. Following
\S\ref{sec:structural-validity}, we compute coefficients of variation
(CVs) across runs for state-node and edge counts, average out-degree,
maximum AXTree depth, and {\tt fill}, {\tt click}, and {\tt choice}
counts. For six page types (homepage, collection, product, policy, cart,
and search), we remove text, URLs, identifiers, presentational nodes, and
sibling order from each accessibility tree, then measure the
total-variation distance from the cross-run semantic-element centroid
(element distribution deviation) and the normalized deviation in maximum
semantic-tree depth. Finally, we compare full-page screenshots of the
six page types using three content-agnostic design/structure LLM judges,
which give similarity scores (from 0 to 1) of $0.785$ (Opus 5), $0.825$
(GPT-5.6), and $0.875$ (Gemini 3.1 Pro), indicating substantial
agreement across reruns. Table~\ref{tab:generation-stability} summarizes
the results.

\begin{table}[h!]
\centering
\caption{Consistency across three generation runs of the same shop
specification.}
\label{tab:generation-stability}
\small
\begin{tabular}{llc}
\toprule
\textbf{Analysis} & \textbf{Metric} & \textbf{Result} \\
\midrule
Website quality & Node count CV & 2.89\% \\
 & Edge count CV & 7.73\% \\
 & Avg.\ out-degree CV & 4.76\% \\
 & Max AXTree depth CV & 7.96\% \\
 & {\tt fill} count CV & 0\% \\
 & {\tt click} count CV & 4.10\% \\
 & {\tt choice} count CV & 0.99\% \\
\midrule
Website structure & Element distribution deviation & 0.109 \\
 & Maximum depth deviation & 0.086 \\
\midrule
Visual comparison & Mean LLM-judge visual similarity & 0.828 \\
\bottomrule
\end{tabular}
\end{table}

\textbf{Cross-model comparison.} We also evaluate a shop generated by
GLM-5.2 (Appendix~\ref{app:glm-results}) from the same specification;
structural distances are recomputed against the joint four-shop
centroid (Table~\ref{tab:cross-model-stability}). The GLM-generated shop
retains the shared semantic structure and broad design language, with
moderate differences concentrated in the cart and search pages.

\begin{table}[h!]
\centering
\caption{Consistency of the three GPT-5.5-generated shops and one
GLM-5.2-generated shop from the same specification, measured against
the joint four-shop centroid.}
\label{tab:cross-model-stability}
\small
\begin{tabular}{lcc}
\toprule
\textbf{Metric} & \textbf{GPT-5.5 shops} & \textbf{GLM-5.2 shop} \\
\midrule
Element distribution deviation & 0.118 & 0.107 \\
Maximum depth deviation & 0.097 & 0.145 \\
Mean visual similarity & 0.886 & 0.620 \\
\bottomrule
\end{tabular}
\end{table}

\clearpage

\section{Shop Generation with an Open-Weight Model}
\label{app:glm-results}

ShopGym treats the coding-agent harness and its underlying model as
replaceable components rather than fixed dependencies, so the cost
figures in \S\ref{sec:scalability} characterize one instantiation (the
Pi harness driving GPT-5.5) rather than the pipeline itself. To check
that the pipeline is not tied to a single frontier provider, we
additionally run one shop-generation job with GLM-5.2, holding the
harness and all pipeline stages fixed.

The GLM-5.2 run completes the pipeline and produces a buildable,
servable storefront with comparable website-structure metrics
(Appendix~\ref{app:generation-stability}).
Table~\ref{tab:glm-generation-cost} reports its token usage, wall-clock
time, and approximate API cost alongside the corresponding GPT-5.5
shop-generation stage. The GLM-5.2 run costs $82\%$ less while consuming
roughly twice the tokens and about $1.5$--$2\times$ the wall-clock time.
This demonstrates provider flexibility and lower API cost, not lower
compute.

\begin{table}[h!]
\centering
\caption{Shop-generation stage cost under GPT-5.5 and GLM-5.2, both
driven by the Pi harness. The GPT-5.5 row aggregates six runs (three
runs on each of two shops); the GLM-5.2 row is a single
shop-generation run. Token counts include cached input, uncached
input, and output.}
\label{tab:glm-generation-cost}
\small
\begin{tabular}{lccc}
\toprule
\textbf{Model} & \textbf{Token usage} & \textbf{Wall-clock time} & \textbf{Approx.\ API cost} \\
\midrule
GPT-5.5 & 17--18M & 2--3 h & \$46--48 \\
GLM-5.2 & 37.8M   & 4.5 h  & \$8.47 \\
\bottomrule
\end{tabular}
\end{table}

\clearpage

\section{Diversity Across Generated Shops}
\label{app:shop-diversity}

This appendix expands \S\ref{sec:scalability}. Scale is only useful if
additional shops are meaningfully different from one another. Across
the three {\tt sandbox shops} and the seven additional fully synthetic
shops, we inspect the homepage, one collection page, and one product
page per shop. Table~\ref{tab:shop-diversity} shows that the shops span
a wide range on every interface-complexity measure: homepage DOM size
varies by roughly a factor of six, visible interactive homepage elements
by a factor of seven, and collection-page filtering controls range from
none to $93$ checkboxes. The variation is therefore in interaction
density and information architecture, not merely in color and branding.

\begin{table}[h!]
\centering
\caption{Range of interface-complexity measures across the ten
generated shops (homepage, one collection page, and one product page
inspected per shop).}
\label{tab:shop-diversity}
\small
\begin{tabular}{lcc}
\toprule
\textbf{Environment measure} & \textbf{Minimum} & \textbf{Maximum} \\
\midrule
Homepage DOM elements                  & 238 & 1{,}462 \\
Visible interactive homepage elements  & 24  & 164 \\
Interactive elements before scrolling  & 9   & 54 \\
Homepage buttons                       & 3   & 57 \\
Homepage images                        & 4   & 123 \\
Interactive collection-page elements   & 48  & 249 \\
Collection-page checkboxes             & 0   & 93 \\
\bottomrule
\end{tabular}
\end{table}

Common shopping operations are also realized through different
interface forms across shops, so an agent cannot rely on a single
storefront template. Table~\ref{tab:shop-interface-forms} lists the
observed forms and how many of the ten shops expose each; coverage
counts overlap because a shop may expose several forms for the same
operation. Thus, shops share core e-commerce operations while varying
interaction density, information architecture, control representation,
configuration structure, and domain-specific information requirements.
These measured ranges establish structural and interface variation
across the generated shops; they are not by themselves proof that
distinct latent capabilities are required.

\begin{table}[h!]
\centering
\caption{Interface forms through which each shopping operation is
exposed, and the number of the ten generated shops that expose each
form. Counts overlap because a single shop may expose several forms for
the same operation.}
\label{tab:shop-interface-forms}
\small
\begin{tabular}{p{0.22\linewidth} p{0.70\linewidth}}
\toprule
\textbf{Skill} & \textbf{Observed interface forms and coverage} \\
\midrule
Product discovery & collection links (10/10), direct homepage product links (7/10), category-first navigation (3/10), carousel controls (6/10) \\
\addlinespace[2pt]
Search & icon or button (9/10), visible search box (4/10), direct link (4/10) \\
\addlinespace[2pt]
Filtering & checkboxes (6/10), filter panel (4/10), price controls (3/10), no explicit filters (2/10) \\
\addlinespace[2pt]
Sorting & standard dropdown (6/10), shop-specific menu (3/10), combined filter/sort panel (1/10), no explicit sorting (1/10) \\
\addlinespace[2pt]
Product options & dropdowns (2/10), single-choice controls (3/10), buttons or color samples (5/10), personalization grid (1/10) \\
\addlinespace[2pt]
Cart & homepage link (6/10), header button or icon (4/10), product-page add action (10/10), bundle action (1/10) \\
\addlinespace[2pt]
Product/policy details & policy links, direct details, and tabbed or click-to-open sections appear across all 10 shops \\
\bottomrule
\end{tabular}
\end{table}

\clearpage

\section{ShopGuru E2E Prompt Templates}
\label{app:e2e-prompts}

This appendix accompanies \S\ref{sec:shopguru-e2e} and reproduces (i)
the system prompt used by the LLM-authored long-horizon journey
generator, (ii) the user prompt template skeleton (placeholders only),
and (iii) three of the eight few-shot examples that ground the LLM in
human-shopper behavior. The selected examples are the ones that
best span the behavior categories enumerated in
Table~\ref{tab:shopguru-skills}: \emph{Multi-pet household shopping}
(multi-product cart with a quantity edit), \emph{Free-shipping
threshold calculation} (cart-state arithmetic with a localized
banner), and \emph{Filter that returns zero results, then recover}
(filter-recover). The remaining five few-shot examples are available
in the released benchmark code. All prompt text is reproduced
verbatim modulo ASCII-folding of a small number of non-ASCII glyphs
($\rightarrow$, currency symbols, accented Lithuanian characters).

\subsection{System Prompt}
\label{app:e2e-system-prompt}

\begin{lstlisting}[style=shopprompt]
You are an expert at designing evaluation tasks for e-commerce web agents.
Your job is to author realistic end-to-end shopping journeys that mirror how
real human buyers actually browse and transact on online stores.

Hard rules:
- Every task must be 100%
  products, collections, page titles, or variant options that are not in
  the data dump below.
- A product mentioned in a task must exist in the catalog (Title + handle).
- A collection mentioned in a task must exist (Title + handle).
- If you ask the agent to "select a Color variant" or "select a Size",
  the named product MUST have that variant option AND values. Otherwise
  use generic phrasing like "select any variant" or simply "add it to cart".
- A product/collection pairing in the same task (e.g. "Navigate to X
  collection and add product Y") is only valid if Y is actually a member
  of X according to the Product-to-Collection Mappings section.
- Every task must end in a well-defined state the agent can reach: a
  specific set of items in the cart, a specific URL visited for a
  navigation task. Never "and then browse around".
- **Never instruct the agent to click Checkout, proceed to checkout, head
  to a checkout page, abandon checkout, or otherwise traverse the cart ->
  checkout boundary.** Checkout is intentionally disabled on the sandbox
  shop and the LLM judge grades tasks on the **final cart state only**.
  End shopping tasks with wording like "Once the product is in your cart,
  end the session. Do not click any Checkout button." For tasks involving
  cart edits (add -> remove, quantity change), describe the expected final
  cart explicitly.
- Intents should be second-person, imperative. Shopping tasks should be
  one or two short paragraphs; they may include natural detours (policy
  lookup, About page, brand comparison) before the cart-state terminal.
- Avoid tasks that require user authentication, payment entry, or any
  externally-gated flow.
- Do not produce tasks that require human judgment calls the agent cannot
  verify (e.g., "pick the most stylish product").

Quality bar (this is what separates a great task from a basic one):
- Open with a one-sentence persona or motivation that frames the journey
  ("First-time visitor.", "You're a returns-cautious shopper.",
  "Multi-pet household shopping.", "Sales hunting.").
- Reference specific storefront UI elements (homepage banner, top menu,
  footer Quick Links, brand menu) by name when they exist.
- Mix at least 2-3 distinct skills per task (search + compare + cart edit;
  policy lookup + nav drilldown + add). Single-step "search and add"
  tasks are too easy and should be the minority.
- Use exact brand/product/collection titles as written in the data --
  never paraphrase ("Fresh Drops" not "new arrivals", "Womens Run Club"
  not "running gear").

Output format: a JSON object with a single key "tasks" containing an array
of task objects, each with fields:
  id (slug-friendly string),
  type ("shopping" or "navigation"),
  intent (string),
  success_criteria (object with url_contains and a descriptive type field).
Do not include the `url` field -- that will be filled in downstream.
\end{lstlisting}

\subsection{User Prompt Template Skeleton}
\label{app:e2e-user-template}

The user prompt is built per shop by substituting the placeholders
below with concrete data extracted by ShopArena. Sections marked
\texttt{\{placeholder\}} are filled in at generation time from
\texttt{products.json}, \texttt{collections.json}, \texttt{stats.json},
and \texttt{pages.json}.

\begin{lstlisting}[style=shopprompt]
Author {count} end-to-end evaluation tasks for the following storefront.
Follow the system prompt's schema and quality bar. Cover as many of the
behavior categories below as this store supports, and skip any that are
not applicable to this store.

### Shop profile
Name: {store_name}
Description: {store_description}
Country: {country}, Currency: {currency}, Language: {language}
Domain: {real_url}

### Top collections (title -- handle)
{collections_list}

### Top product types
{product_types_list}

### Example products (title -- handle, with REAL variant options & values)
{products_list}

### Product-to-Collection memberships (use these to ground multi-step tasks)
{mapping_str}

### Collection facets / option patterns (for filter tasks)
{option_patterns_list}

### Policy / info pages (title -- /pages/handle)
{pages_list}

### Gift card products
{gift_card_list_or_none}

### Behavior categories to cover (aim for at least 10 of 14)
1.  Search & atomic add-to-cart
2.  Nav drilldown (menu -> sub-menu -> collection -> product)
3.  Filter + sort (e.g. Format=Hardcover then sort by price)
4.  Filter that returns zero results, then recover
5.  Substitute-match discovery (intended product missing -> close alternative)
6.  Review / detail read on a product page
7.  Size chart / fit guide lookup
8.  Shipping policy lookup (with cart action after)
9.  Returns / refunds lookup (with cart action after)
10. Gift card purchase (only if the shop sells gift cards)
11. Multi-product cart with edit (add A, add B, remove A, set qty 2 on B)
12. Cross-collection or cross-brand comparison (compare A and B, pick one)
13. Contact / store locator / about page
14. Free-shipping threshold or sale-discount calculation (if banner exists)

### High-quality reference tasks (from other shops in this benchmark suite)
[See Appendix sections below for three representative few-shot examples.
Five additional examples are included in the released benchmark code.]

### Anti-patterns to AVOID
- "Search for X. Pick a Color and Size variant. Add to cart." -- too thin,
  no persona, no detour, single skill.
- Mentioning "Color" or "Size" for a product whose options list above does
  NOT include that name.
- Generic UI references like "navigate to the homepage" without naming a
  specific section or CTA.
- Two products in one intent that are not co-located in any collection
  according to the membership list above.

When constructing tasks, USE EXACT product titles, collection titles, and
option names from the lists above. Use the storefront's own wording for
collection names ("Fresh Drops" not "new arrivals"). For multi-step tasks,
ensure each (collection, product) pair you mention is a real membership.

Respond with a JSON object with a "tasks" key containing the array of
{count} tasks. Each `id` should use the format
"{shop_slug}-e2e-v1-{index}" replacing {index} with numbers starting at 1.
\end{lstlisting}

\subsection{Few-Shot Example A: Multi-Pet Household Shopping}
\label{app:e2e-fewshot-multipet}

A multi-product cart with a per-item quantity edit and a cross-collection
nav. Exercises behavior categories 2 (Nav drilldown), 11 (Multi-product
cart with edit), and the cart-state-only termination rule.

\begin{lstlisting}[style=shopprompt]
intent: "Multi-pet household shopping. Navigate to the Dog Kibble
collection via the Dogs menu and add any kibble product to cart.
Navigate to Cats -> Litter from the top menu and open Worlds Best Cat
Litter. Select any size variant (for example Original 28lb) and add it
to cart. Open the full cart page and increase the cat litter quantity
from 1 to 2. Once the cart contains one dog kibble and two units of
cat litter, end the session. Do not click any Checkout button."
success_criteria.type: "cart_multi_pet_with_quantity_edit"
\end{lstlisting}

\subsection{Few-Shot Example B: Free-Shipping Threshold Calculation}
\label{app:e2e-fewshot-freeshipping}

Cart-state arithmetic over a localized homepage banner. Exercises
behavior category 14 (Free-shipping threshold) and demonstrates that
the prompt is robust to non-English storefronts (here, Lithuanian).

\begin{lstlisting}[style=shopprompt]
intent: "Hit the free-shipping threshold. Note the 'Nemokamas
pristatymas nuo 49 EUR' banner on the homepage. Navigate to Sunims ->
Skanestai and add a treat product under EUR 25 to the cart. Open the
cart page, observe how much more you need to reach EUR 49 for free
shipping. Go back to the store, navigate to the same Skanestai
category, and add a second product that pushes the total over EUR 49.
Return to cart, confirm the free-shipping banner has updated. Once
both products are in your cart with a total over EUR 49, end the
session. Do not click any Checkout button."
success_criteria.type: "cart_after_free_shipping_threshold"
\end{lstlisting}

\subsection{Few-Shot Example C: Filter-Recover}
\label{app:e2e-fewshot-filterrecover}

A filter that yields zero or sparse results, followed by a recovery
strategy. Exercises behavior category 4 (Filter that returns zero
results, then recover) and demonstrates handling of edge cases that
deterministic generators cannot synthesize on their own.

\begin{lstlisting}[style=shopprompt]
intent: "Navigate to the Christmas & Hanukkah collection via the Shop
by Holidays & Events menu. Apply a Size filter for '3-6 Months';
results will likely be sparse or empty. Observe the empty or filtered
state, then clear the Size filter and apply a different one such as
Color = Blue. If products appear, open the first one, select any
variant, and add it to cart; if still empty, remove all filters and
pick any product from the unfiltered list. Once a product is in your
cart, end the session. Do not click any Checkout button."
success_criteria.type: "cart_after_filter_recover"
\end{lstlisting}

\end{document}